\documentclass{article}

\PassOptionsToPackage{table}{xcolor}
\usepackage{fancyhdr,natbib}
\usepackage{preprint}
\usepackage{times}

\usepackage{amsmath,amsfonts,bm}

\def\eqref#1{equation~\ref{#1}}

\def\1{\bm{1}}

\DeclareMathAlphabet{\mathsfit}{\encodingdefault}{\sfdefault}{m}{sl}
\SetMathAlphabet{\mathsfit}{bold}{\encodingdefault}{\sfdefault}{bx}{n}

\usepackage[T1]{fontenc}
\usepackage{microtype}
\usepackage[table]{xcolor}
\usepackage[most]{tcolorbox}
\usepackage{xurl}
\usepackage{hyperref}

\usepackage{graphicx}
\usepackage{booktabs}
\usepackage{multirow}
\usepackage{enumitem}
\usepackage{float}
\usepackage{wrapfig}

\usepackage{amsmath}
\usepackage{amssymb}
\usepackage{amsfonts}

\usepackage{algorithm}
\usepackage{algpseudocode}

\definecolor{lightGray}{HTML}{ECECEC}
\definecolor{needleGreen}{HTML}{B6D9C5}
\definecolor{longBlue}{HTML}{BBD9EC}
\definecolor{skyBlue}{HTML}{87CEEB}
\definecolor{lavendar}{HTML}{F6DDF3}
\definecolor{trainGray}{RGB}{242,242,242}
\definecolor{attnBlue}{RGB}{232,241,252}
\definecolor{oursGreen}{RGB}{226,244,233}
\definecolor{mygreen}{rgb}{0.92, 1.0, 0.92}
\definecolor{mygray}{gray}{0.95}
\definecolor{myblue}{rgb}{0.88,0.98,1}
\definecolor{promptTitle}{HTML}{3F4752}
\definecolor{promptFrame}{HTML}{D6DAE0}
\definecolor{promptBody}{HTML}{FBFCFD}
\definecolor{promptTitleBg}{HTML}{EEF1F4}

\tcbset{
  promptbox/.style={
    enhanced,
    breakable,
    colback=promptBody,
    colframe=promptFrame,
    colbacktitle=promptTitleBg,
    coltitle=promptTitle,
    fonttitle=\bfseries\footnotesize,
    fontupper=\footnotesize,
    boxrule=0.45pt,
    arc=2pt,
    left=6pt,
    right=6pt,
    top=6pt,
    bottom=4pt,
    toptitle=3pt,
    bottomtitle=3pt,
    titlerule=0.45pt,
    before skip=4pt,
    after skip=6pt,
  }
}

\title{ReVEAL: Eliciting Role-Aware Latent Visual Evidence from MLLMs for Reasoning Video Object Segmentation}

\author{
  Zhongyuan Bao$^{1}$, Cunjie He$^{1}$, Lejun Zhang$^{2}$, Zenglin Xu$^{1}$, Yanwei Fu$^{1}$ \\
  $^{1}$Fudan University \qquad $^{2}$New York University
}

\hypersetup{
  pdfauthor={Zhongyuan Bao, Cunjie He, Lejun Zhang, Zenglin Xu, Yanwei Fu},
  pdftitle={ReVEAL: Eliciting Role-Aware Latent Visual Evidence from MLLMs for Reasoning Video Object Segmentation}
}

\begin{document}

\maketitle

\begingroup
\renewcommand{\thefootnote}{}
\makeatletter
\long\def\@makefntext#1{\noindent#1}
\makeatother
\footnotetext{Project Page: \url{https://github.com/Lexarymade/ReVEAL}}
\endgroup

\begin{abstract}
Reasoning video object segmentation (ReVOS) requires segmenting the target specified by an implicit query throughout a video.
Existing MLLM-based methods typically require costly cross-model alignment, tying MLLM training to a downstream segmentation model.
We present \textbf{ReVEAL}, a decoupled \textbf{thinking-with-video} framework for ReVOS that elicits role-aware latent visual evidence from MLLM reasoning to directly guide segmentation.
ReVEAL introduces \textbf{Visual Evidence Tokens} (VETs) into video reasoning to associate necessary visual regions or entities with target, background, or distractor roles, while a Dynamic Attention Layer Router adaptively aggregates VET-to-video attention across layers into role-specific grounding evidence.
We train the MLLM to perform VET-augmented reasoning on our constructed ReVEAL-47K dataset, with a Contrastive Attention Grounding Loss as auxiliary supervision to apply role-specific constraints to VET attention and optimize layer routing for reliable latent grounding.
At inference, ReVEAL identifies the target through this thinking-with-video process and fuses the resulting role-specific evidence to guide a frozen SAM2 in predicting target masks.
Experiments on five benchmarks show that ReVEAL outperforms alignment-free baselines and remains competitive with alignment-based methods.

\end{abstract}

\section{Introduction}
\label{sec:intro}


Reasoning Video Object Segmentation (ReVOS) aims to segment target objects throughout a video according to free-form and often implicit language expressions. Unlike conventional referring video object segmentation (RVOS), the target may not be identifiable from a single frame or an explicit category name. Instead, it relies on the model's ability to perceive and understand the video content in relation to the query, such as identifying \emph{``the object used to keep the drink cold''}. This makes Multimodal Large Language Models (MLLMs)~\citep{liu2023llava,bai2023qwenvl} a natural backbone for ReVOS. However, the challenge is not only to understand the expression, but also to connect such understanding to the visual evidence that supports a consistent object mask across time.

Since MLLMs do not directly produce pixel masks, existing MLLM-based ReVOS methods typically rely on cross-modal alignment with an external segmenter, making MLLM training dependent on the downstream segmentation model. Existing approaches often introduce a dedicated segmentation token, such as \texttt{<SEG>}, and align its hidden representation with the prompt space of an external segmenter~\citep{kirillov2023sam,ravi2024sam2}.
Although straightforward for mask generation, such design incurs additional training costs and ties the framework to a specific MLLM–segmenter pair, complicating adaptation to alternative backbones and downstream settings.

Recent reasoning MLLMs adopt a thinking-with-images paradigm that incorporates visual primitives into intermediate chain-of-thoughts (CoT) steps, using boxes and points~\citep{openai-o3,segzero,treebench} or special tokens~\citep{qin2025chain,padt,samtok}. This paradigm extends to thinking-with-video, where spatio-temporal cues serve as intermediate references for reasoning across frames~\citep{zhang2025vtimecot,meng2025open}. Interactive video segmentation models such as SAM2~\citep{ravi2024sam2} likewise accept spatial primitives as prompts for object identification and mask propagation, providing a natural interface through which visual evidence produced during MLLM reasoning can directly guide segmentation~\citep{kao2025cot}.

Yet these primitives are typically limited to selected keyframes~\citep{veasonr1,omnir1}, providing only sparse grounding evidence across the video. Causal self-attention, on the other hand, offers a denser alternative~\citep{vaswani2017attention}: text tokens attend to visual tokens across preceding frames, yielding video-wide attention maps with spatial grounding cues~\citep{tam}, although these cues may not reliably localize the corresponding regions~\citep{trag}.
DecAF~\citep{decaf} relies on a tailored multi-rollout procedure to obtain object and background attention maps for segmentation, incurring substantial inference cost without leveraging video reasoning itself.
This motivates us to explore whether an MLLM can produce the grounding evidence needed for segmentation as part of its own video reasoning within a single response.

\begin{figure*}[t]
  \centering
  \includegraphics[width=\textwidth]{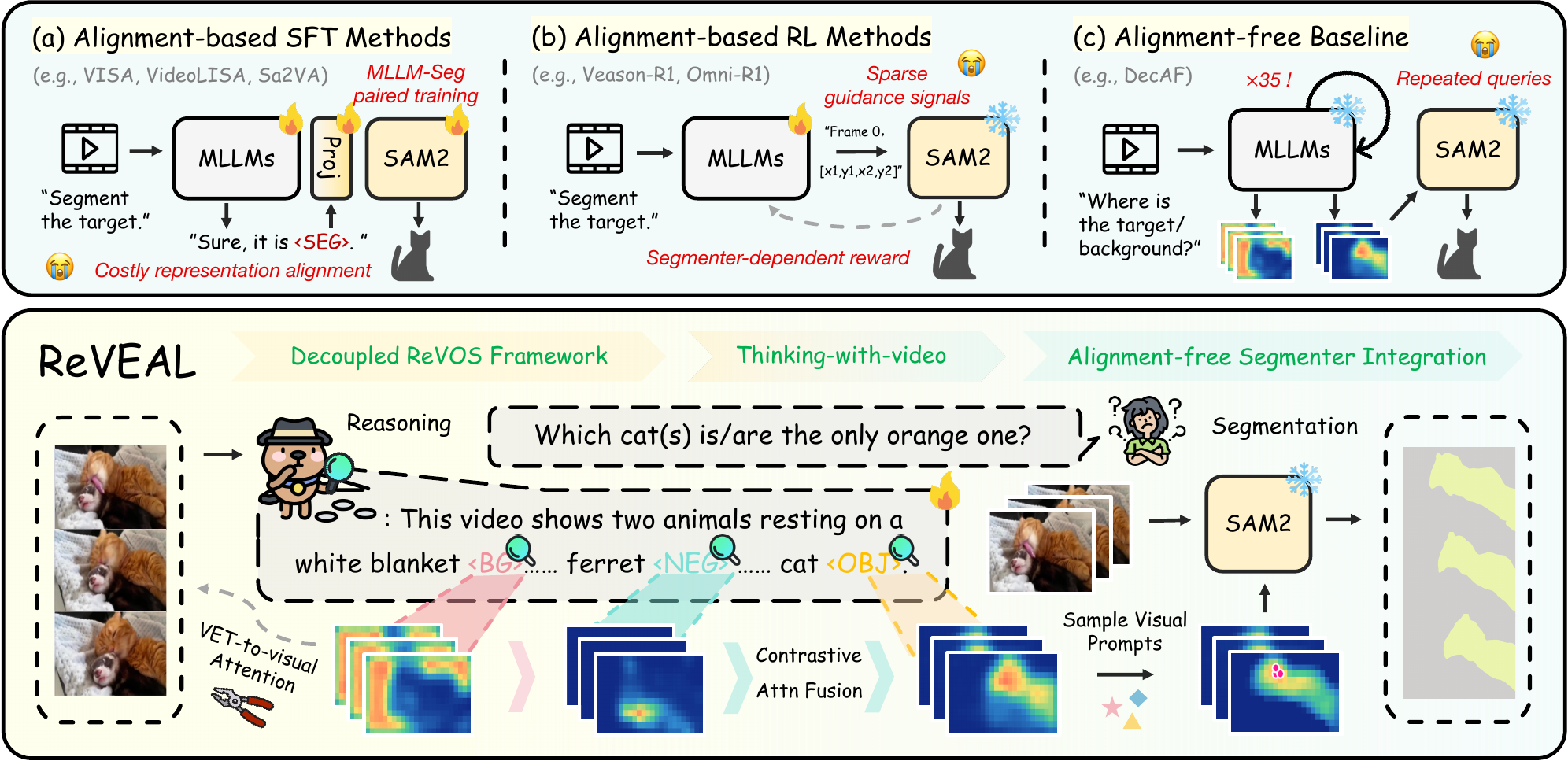}
  \setlength{\abovecaptionskip}{0pt}
  \caption{\textbf{Overview of ReVEAL.} Previous methods mainly involve the segmenter during training or rely on multiple MLLM queries at inference to obtain grounding signals. In contrast, ReVEAL adopts a decoupled ReVOS framework that elicits internal visual evidence from the MLLM’s own video reasoning, enabling alignment-free integration with an external segmenter.}
  \label{fig:paradigm}
\end{figure*}

To this end, we present \textbf{ReVEAL}, a thinking-with-video framework for ReVOS that elicits role-aware latent visual evidence from MLLM reasoning to guide segmentation.
ReVEAL introduces Visual Evidence Tokens (VETs) to mark necessary visual regions or entities as target (\texttt{<OBJ>}), background (\texttt{<BG>}), or distractors (\texttt{<NEG>}) during reasoning.
A Dynamic Attention Layer Router (DALR) learns to adaptively aggregate their attention over preceding visual tokens across Transformer layers into role-specific evidence maps. 
%
To enable the MLLM to elicit faithful role-aware latent visual evidence through video reasoning, we perform supervised fine-tuning (SFT) for VET-augmented reasoning on our constructed ReVEAL-47K dataset, using a Contrastive Attention Grounding Loss (CAGL) as auxiliary supervision to optimize layer routing and apply role-specific constraints to VET attention over video tokens for reliable latent grounding.

At inference, ReVEAL identifies the target through VET-augmented video reasoning. DALR then aggregates VET-to-video attention from the CoT across Transformer layers into role-specific evidence.
These attention maps are then contrastively fused, and point prompts are sampled from high-confidence regions to guide a frozen external SAM2 in predicting the target masks, forming an alignment-free interface between MLLM reasoning and video object segmentation.

In a nutshell, our contributions are summarized as follows:


\begin{itemize}[leftmargin=1.2em,itemsep=2pt,topsep=2pt]

\item We present \textbf{ReVEAL}, a decoupled thinking-with-video framework that elicits role-aware latent visual evidence during MLLM reasoning to guide segmentation without cross-model alignment.

\item We equip the MLLM with this capability by combining reasoning supervision on \textbf{ReVEAL-47K} with \textbf{CAGL} for attention grounding over layers adaptively selected and weighted by \textbf{DALR}. 

\item We demonstrate that internal signals from MLLM reasoning can provide \textbf{effective grounding cues} for video object segmentation through experiments on five RVOS and ReVOS benchmarks.

\end{itemize}

\section{Related Work}
\label{sec:related}

\subsection{Multimodal Large Language Models}

Building on the success of LLMs in natural language processing, image MLLMs extend language models to visual inputs by bridging visual features into the language model through cross-attention resamplers~\citep{alayrac2022flamingo}, querying transformers~\citep{li2023blip2,dai2023instructblip}, or projection-based visual instruction tuning~\citep{liu2023llava,zhu2023minigpt4}. Subsequent systems further scale visual backbones and training data to improve general vision--language capabilities~\citep{chen2024internvl,bai2023qwenvl,li2024llavaonevision,wang2024qwen2vl}. Video MLLMs extend this paradigm to video by incorporating sampled-frame representations and temporal modeling~\citep{zhang2023videollama,maaz2024video,lin2024videollava,li2026videochat3}. Beyond general video understanding, recent work increasingly adopts video MLLMs as general-purpose backbones for downstream tasks that require fine-grained perception and reasoning, including dense video captioning~\citep{sharegpt4video,xu2024pllava}, temporal grounding~\citep{ren2024timechat,huang2024vtimellm}, subject-specific video question answering~\citep{yuan2025videorefer,dam}, and embodied spatial reasoning~\citep{vsi}. 
We extend this line of research to ReVOS by leveraging MLLMs' spatio-temporal reasoning capabilities to identify target objects specified by complex queries.


\subsection{Referring and Reasoning Video Object Segmentation}

Referring Video Object Segmentation (RVOS) segments an object in a video based on a natural language expression. Early methods focus on explicit expressions through cross-modal fusion or query propagation~\citep{ytvos,botach2022mttr,wu2022referformer}. Reasoning Video Object Segmentation (ReVOS) extends this task to implicit queries that identify targets through their functions, behaviors, or relations to other objects, requiring more complex reasoning~\citep{yan2024visa}. 
Most MLLM-based methods interpret implicit queries and learn cross-modal alignment by mapping the hidden state of a special \texttt{<SEG>} token to a downstream segmenter's prompt space for mask prediction~\citep{lai2024lisa,yan2024visa,bai2024videolisa,interrvos,sa2va,glus,vrshq}.
Recent work promotes this alignment indirectly through reinforcement learning, using SAM2-based segmentation rewards to optimize key-frame selection and MLLM-generated bounding boxes~\citep{veasonr1,omnir1}.
Both strategies tie MLLM training to a downstream segmenter, leaving the potential of native MLLM reasoning as an independent source of visual grounding evidence underexplored~\citep{kao2025cot}.
In contrast, we train the MLLM under a thinking-with-video paradigm to elicit latent grounding evidence from video reasoning and use this evidence to guide an external segmenter at inference.

\subsection{Visual Evidence in MLLM Reasoning}
OpenAI-o3~\citep{openai-o3} introduces a thinking-with-images paradigm that incorporates references to visual regions into intermediate reasoning steps. Recent methods explicitly refer to these regions using boxes and points~\citep{segzero,treebench} or special tokens representing masks and image patches~\citep{qin2025chain,padt,vrt}.
This idea also extends to thinking-with-video, where spatio-temporal cues serve as references for video reasoning~\citep{meng2025open}. 
However, generating such references throughout a video becomes cumbersome for dense video prediction. Instead, attention from output tokens to the preceding multimodal context provides a potential source of latent grounding evidence by highlighting visual regions associated with token semantics~\citep{kang2025your,wu2024flmm,tam,hu2026attend,vaswani2017attention}. 
Despite this potential, raw attention is often noisy and susceptible to attention sinks~\citep{huang2024opera,darcet2024vision}, limiting its reliability as visual evidence.
InterRVOS~\citep{interrvos} constrains attention to relevant object regions to enrich semantic representations during cross-model alignment, whereas DecAF~\citep{decaf} aggregates attention maps across multiple queries for visual grounding. 
In contrast, ReVEAL derives role-aware latent visual evidence directly from video reasoning and fuses those cues for more accurate target localization.
\section{Method}
\label{sec:method}

Reasoning video object segmentation (ReVOS) aims to segment the target specified by an implicit query throughout a video. We propose ReVEAL, a decoupled thinking-with-video framework that uses role-aware latent evidence from MLLM reasoning to guide segmentation. 
During training, the MLLM learns to use VETs as role-specific semantic anchors for necessary visual regions or entities, with spatio-temporal grounding constraints applied directly to its own attention rather than through alignment with a downstream segmenter.
At inference, ReVEAL derives role-specific latent visual evidence through VET-augmented video reasoning and fuses this evidence as grounding signals to guide a frozen SAM2 in predicting target masks.

\begin{figure}[t]
  \centering
  \includegraphics[width=\linewidth]{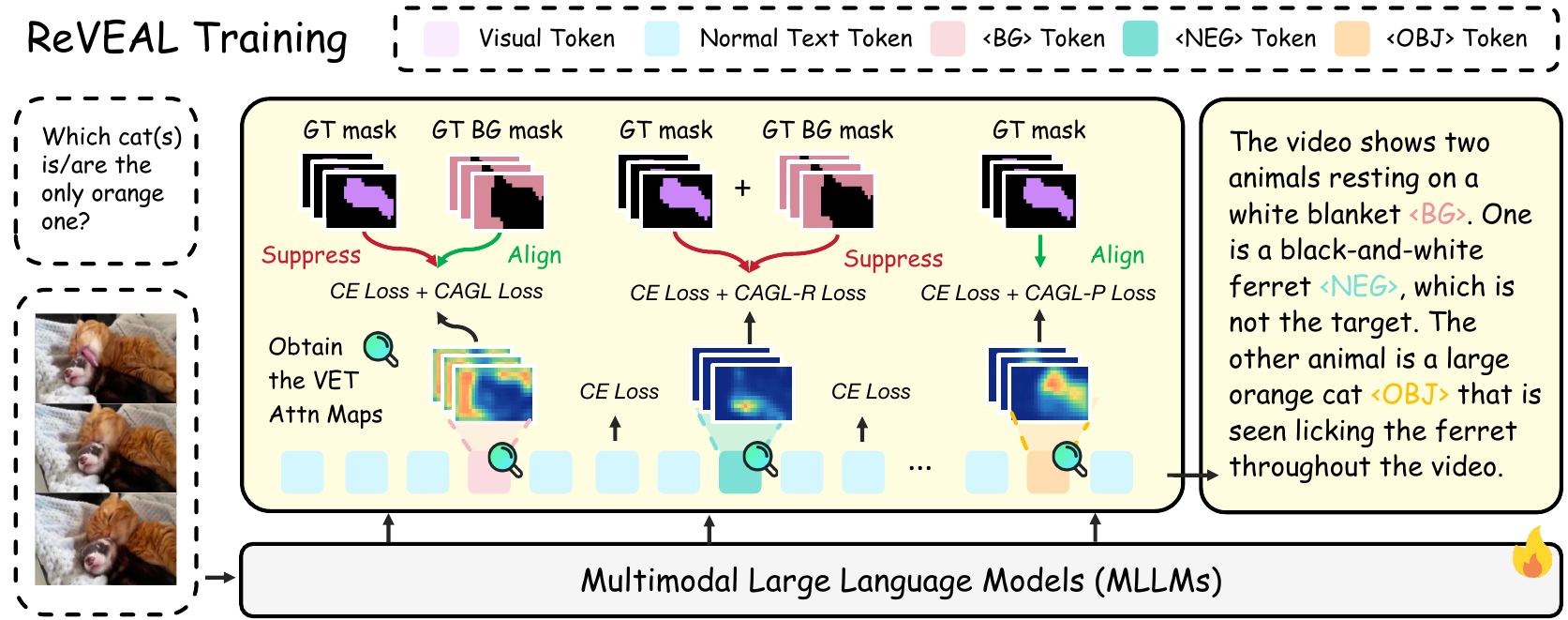}
  \setlength{\abovecaptionskip}{0pt}
  \caption{\textbf{Training of ReVEAL.} We train the MLLM to perform VET-augmented video reasoning using cross-entropy loss, with CAGL providing auxiliary role-specific constraints on VET-to-video attention for reliable latent grounding. }
  \label{fig:overview}
\end{figure}

\paragraph{Taxonomy of VETs.}

ReVEAL augments the MLLM vocabulary $\mathcal{V}_{\mathrm{orig}}$ with three role-specific VETs: \texttt{<OBJ>} specifies the referred target, \texttt{<NEG>} identifies the foreground distractor to exclude, and \texttt{<BG>} indicates the background region to suppress:
\begin{equation}
  \mathcal{V}=\mathcal{V}_{\mathrm{orig}}\cup\mathcal{V}_{\mathrm{VET}}
  \qquad 
  \mathcal{V}_{\mathrm{VET}}= \{\texttt{<OBJ>},\texttt{<BG>},\texttt{<NEG>}\}
\end{equation}


\begin{wrapfigure}{r}{0.43\textwidth}
  \centering
  \vspace{-\intextsep}
  \includegraphics[width=0.41\textwidth]{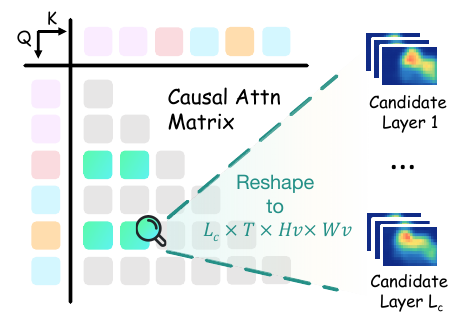}
  \vspace{2pt}
  \setlength{\abovecaptionskip}{0pt}
  \caption{VET Attention Extraction.}
  \label{fig:t2v}
\end{wrapfigure}

\paragraph{Obtaining VET Attention Maps.}

We treat each VET's attention to the preceding visual tokens as a role-specific latent grounding cue. The corresponding attention map is obtained by aggregating these cues across Transformer layers, with later layers selected as candidates because they generally encode stronger semantic signals~\citep{decaf}.
%
%
Let $\mathcal{L}_c$ denote the set of $L_c$ candidate layers and $A\in[0,1]^{L_c\times N\times N}$ collect their post-softmax attention weights averaged across heads, where $N$ is the whole sequence length. For the $j$-th VET occurrence with role $r$, we extract its attention weights over the preceding video tokens and reshape them into layer-wise attention maps $a^{r,j}\in[0,1]^{L_c\times T\times H_v\times W_v}$, where $T$ is the number of frames and $H_v$ and $W_v$ denote the height and width of each frame’s visual token grid.

However, the layers most relevant to target grounding may vary across different cases~\citep{kang2025your}. Inspired by mixture-of-experts routing~\citep{deepseekv3}, we introduce a Dynamic Attention Layer Router (DALR) to dynamically select and weight candidate layers based on the \texttt{<OBJ>} attention maps.
Given the \texttt{<OBJ>} attention maps for each video-query pair, DALR scores the candidate layers, selects the top-$k$ layers as $\mathcal{S}\subset\mathcal{L}_c$, and normalizes their scores into routing weights $w_l$. These weights determine the contribution of each selected layer when aggregating VET attention during both training and inference. More details are provided in Appendix~\ref{app:details}.

\subsection{Learning Video Reasoning with VETs}
\label{sec:method:approach}

\paragraph{Supervising Thinking-with-video Reasoning.}


Through SFT, the MLLM learns to identify query-relevant visual entities or regions during video reasoning and place the corresponding role-specific VETs after their names or descriptions, conditioning each VET occurrence on an entity-specific context.
Given a video $V$, a query $q$, and a VET-augmented response $\mathbf{u}=(u_1,\ldots,u_M)$, we optimize the model using the standard next token prediction objective:

\begin{equation}
\mathcal{L}_{\mathrm{SFT}}
=
-\sum_{i=1}^{M}
\log p\!\left(u_i \mid u_{<i}, V, q\right).
\end{equation}

Let $n_r(u)$ denote the number of occurrences of role $r$ in response $u$. Each response contains $n_{\texttt{<OBJ>}}(u)=1$ and $n_{\texttt{<BG>}}(u)\in\{0,1\}$, while $n_{\texttt{<NEG>}}(u)$ varies with the number of distinct distractors described in the reasoning trace.



\paragraph{Aligning Attention Maps with Visual Regions.}

Although SFT teaches the model to perform VET-augmented video reasoning to identify relevant visual entities and their semantic roles, it does not ensure that VET attention provides reliable latent grounding evidence. Inspired by AML~\citep{interrvos}, we introduce the Contrastive Attention Grounding Loss (CAGL) to supervise the spatio-temporal distribution of VET attention using complementary positive and repulsive terms.

\begin{equation}
  \mathcal{L}_{\mathrm{CAGL}}
  =
  \sum_{l\in\mathcal{S}}
  w_l
  \left(
  \mathcal{L}_{\mathrm{CAGL-P}}^{l}
  +
  \lambda_{\mathrm{rep}}
  \mathcal{L}_{\mathrm{CAGL-R}}^{l}
  \right).
\end{equation}

Here, $w_l$ is the routing weight assigned by DALR to layer $l$, and $\lambda_{\mathrm{rep}}$ balances the repulsive term.

For each occurrence $j$ of a role $r\in\mathcal{P}=\{\texttt{<OBJ>},\texttt{<BG>}\}$, the positive term applies BCE to align its attention distribution with the corresponding ground-truth (GT) region $m^{r}\in[0,1]^{T\times H_v\times W_v}$:


\begin{equation}
  \mathcal{L}_{\mathrm{CAGL-P}}^{l}
  =
  \sum_{r\in\mathcal{P}}
  \sum_{j=1}^{n_r(u)}
  \operatorname{BCE}\!\left(a_l^{r,j},m^r\right),
\end{equation}

During reasoning, the attention maps of the suppressive roles $r\in\mathcal{R}=\{\texttt{<BG>},\texttt{<NEG>}\}$ provide suppressive evidence, but leakage into unintended regions may cause erroneous suppression. We therefore apply a repulsive term that penalizes the normalized attention mass within a role-specific forbidden region $f^r$.
Specifically, we define
$f^{\texttt{<BG>}}=m^{\texttt{<OBJ>}}$ and
$f^{\texttt{<NEG>}}=m^{\texttt{<OBJ>}}\cup m^{\texttt{<BG>}}$:

\begin{equation}
  \hat{a}_l^{r,j}
  =
  \frac{a_l^{r,j}}
  {\lVert a_l^{r,j}\rVert_1+\epsilon},
  \qquad
  \mathcal{L}_{\mathrm{CAGL-R}}^{l}
  =
  \sum_{r\in\mathcal{R}}
  \sum_{j=1}^{n_r(u)}
  -\log\left(
  1-
  \left\langle
  \hat{a}_l^{r,j},
  f^r
  \right\rangle
  +\epsilon
  \right).
\end{equation}

Here, $\lVert a_l^{r,j}\rVert_1$ sums the attention over all spatio-temporal locations, and the inner product $\langle\hat{a}_l^{r,j},f^r\rangle$ measures the normalized attention mass within the forbidden region.
%

\paragraph{Final Training Objective.}
The final objective combines SFT with CAGL:
\begin{equation}
  \mathcal{L} =
  \mathcal{L}_{\mathrm{SFT}}
  + \lambda_{\mathrm{CAGL}}\,\mathcal{L}_{\mathrm{CAGL}}
\end{equation}
Here, $\lambda_{\mathrm{CAGL}}$ controls the strength of CAGL.

\paragraph{Training Data Construction.}

To provide supervision for ReVEAL, we construct \textbf{ReVEAL-47K}, a training dataset comprising VET-augmented CoT traces and region-level mask annotations, from multiple training sources~\citep{lai2024lisa,davis,ytvos,ding2023mevis,groundmore,yan2024visa}. For each sample, we use GAR~\citep{wang2025grasp} with the GT mask on a representative frame to obtain a target caption. Conditioned on this caption, the query, and the video frames, Gemini 2.5 Pro~\citep{comanici2025gemini} generates a CoT response and inserts VETs at the corresponding positions. Qwen3VL~\citep{bai2025qwen3} is used to verify and filter out unreliable samples. The mask annotations include the retained GT mask and a background mask, obtained by taking the complement of the union of the foreground regions predicted by BiRefNet~\citep{zheng2024birefnet} and the target region. More details are provided in Appendix~\ref{app:dataset}.

\subsection{From VET Attention Maps to Segmentation Masks}
\label{sec:method:fusion}

At inference, ReVEAL identifies the target through VET-augmented video reasoning. DALR adaptively selects and weights Transformer layers to aggregate VET attention into role-specific latent evidence maps, which are then fused to guide mask prediction with a frozen, external SAM2.

Specifically, the \texttt{<OBJ>} attention is aggregated across the selected layers to form target evidence, whereas attention associated with all \texttt{<BG>} and \texttt{<NEG>} tokens is aggregated as suppressive evidence:
\begin{equation}
\begin{aligned}
a^{\mathrm{obj}}
&=
\sum_{l\in\mathcal{S}}
w_l a_l^{\texttt{<OBJ>},1},
\
a^{\mathrm{sup}}
&=
\sum_{r\in\mathcal{R}}
\alpha_r
\sum_{j=1}^{n_r(u)}
\sum_{l\in\mathcal{S}}
w_l a_l^{r,j},
\end{aligned}
\end{equation}
where $\alpha_r$ controls the contribution of each suppressive role.

\paragraph{Contrastive Attention Fusion.}
We fuse the target and suppressive evidence frame by frame. For frame $t$, the corresponding slices $a_t^{\mathrm{obj}},a_t^{\mathrm{sup}}\in\mathbb{R}^{H_v\times W_v}$ are combined by subtracting the suppressive evidence from the target evidence:
\begin{equation}
  \tilde{P}_t
  = \max\!\left(a_t^{\texttt{obj}}-a_t^{\mathrm{sup}},\,0\right)
  \qquad
  P_t
  = \frac{\tilde{P}_t - \min(\tilde{P}_t)}
           {\max(\tilde{P}_t) - \min(\tilde{P}_t) + \epsilon}
\end{equation}
Here, the minimum and maximum are taken over spatial locations, and $\epsilon$ ensures numerical stability. We then sample points from the fused attention map $P_t$ as visual prompts for the frozen SAM2.

\paragraph{SAM2 Prompting and Mask Generation.}

We apply a threshold $\tau_{\mathrm{pq}}$ to the fused attention map and sample points from the retained regions according to their attention strengths as prompts for SAM2 to generate candidate mask tracklets. Following the SAM2 prompting pipeline of DecAF~\citep{decaf}, we remove redundant candidates using non-maximum suppression with an IoU threshold $\tau_{\mathrm{nms}}$, ranking them by point-attention and SAM2 confidence scores. We then filter the remaining tracklets by their temporal consistency with the retained attention regions using $\tau_{\mathrm{trk}}$ and propagate them across all frames to produce the final segmentation masks.

\section{Experiments}
\label{sec:experiments}

\paragraph{Datasets and metrics.}

We evaluate ReVEAL on five RVOS and ReVOS benchmarks: Ref-DAVIS~\citep{davis}, Ref-YouTube-VOS~\citep{ytvos},  MeViS~\citep{ding2023mevis}, ReVOS~\citep{yan2024visa}, and ReasonVOS~\citep{bai2024videolisa}. Following standard practice, we report mask quality using region similarity $\mathcal{J}$, boundary accuracy $\mathcal{F}$, and their average $\mathcal{J}\&\mathcal{F}$.

\paragraph{Implementation Details.}
ReVEAL uniformly samples six frames from each video as input. We finetune the MLLM using LLaMA-Factory~\citep{llamafactory}, with the CAGL weight $\lambda_{\mathrm{CAGL}}$ set to 0.2 and the $\lambda_{\mathrm{rep}}$ in CAGL set to 0.1. The router top-$k$ is set to 6.  
For inference and evaluation, we use SAM2-Hiera-Large~\citep{ravi2024sam2} as the frozen segmentation model.
We set $\alpha_{\mathrm{<BG>}}=0.5$ for Qwen2.5-VL and $\alpha_{\mathrm{<BG>}}=0.1$ for Qwen3VL, while using $\alpha_{\mathrm{<NEG>}}=1.0$ for both.
All other settings follow the baseline~\citep{decaf}, using a threshold of $\tau_{\mathrm{pq}}=0.8$, a NMS IoU threshold of $\tau_{\mathrm{nms}}=0.7$, and a tracklet-consistency threshold of $\tau_{\mathrm{trk}}=0.8$.
Our experiments are conducted on NVIDIA H800 GPUs. More details are provided in Appendix~\ref{app:details}.

\subsection{Main Results}
\label{sec:experiments:main}

\paragraph{Masks with Attn.} 

Table~\ref{tab:attn_mask} compares ReVEAL with attention-derived grounding methods under the same Qwen2.5-VL-7B backbone and six frame input setting. ReVEAL achieves the best performance across all benchmarks, outperforming DecAF by 5.7\%, 3.0\%, and 4.6\% $\mathcal{J}\&\mathcal{F}$ on Ref-DAVIS, ReasonVOS, and ReVOS.
These results indicate that ReVEAL produces attention maps with more complete spatio-temporal coverage of the target regions, providing stronger grounding evidence. Despite these gains, $\mathcal{F}$ remains relatively low because the coarse attention maps lack precise object boundaries and therefore still require a dedicated segmenter for accurate mask prediction.

\begin{table*}[t!]
\centering
\setlength{\tabcolsep}{3pt}
\centering
\caption{Comparison of MLLM-based ReVOS methods using attention-derived masks.}
\label{tab:attn_mask}
\vspace{0.5mm}
\resizebox{\linewidth}{!}{
\begin{tabular}{l l ccc ccc ccc ccc ccc}
\toprule
\multirow{2}{*}{Method} & \multirow{2}{*}{MLLM} &
\multicolumn{3}{c}{Ref-DAVIS} &
\multicolumn{3}{c}{ReasonVOS} &
\multicolumn{3}{c}{ReVOS {\scriptsize (Overall)}} &
\multicolumn{3}{c}{ReVOS {\scriptsize (Referring)}} &
\multicolumn{3}{c}{ReVOS {\scriptsize (Reasoning)}} \\
\cmidrule(lr){3-5}\cmidrule(lr){6-8}\cmidrule(lr){9-11}
\cmidrule(lr){12-14}\cmidrule(lr){15-17}
& & $\mathcal{J}$\&$\mathcal{F}$ & $\mathcal{J}$ & $\mathcal{F}$ &
$\mathcal{J}$\&$\mathcal{F}$ & $\mathcal{J}$ & $\mathcal{F}$ &
$\mathcal{J}$\&$\mathcal{F}$ & $\mathcal{J}$ & $\mathcal{F}$ &
$\mathcal{J}$\&$\mathcal{F}$ & $\mathcal{J}$ & $\mathcal{F}$ &
$\mathcal{J}$\&$\mathcal{F}$ & $\mathcal{J}$ & $\mathcal{F}$ \\
\midrule
\midrule
TAM~\citep{tam} & Qwen2.5VL-7B &
2.7 & 1.8 & 3.7 & 3.3 & 2.7 & 3.8 & 3.0 & 2.8 & 3.1 & 3.0 & 2.9 & 3.2 & 2.9 & 2.7 & 3.0 \\
Loc-Head~\citep{kang2025your} & Qwen2.5VL-7B &
13.0 & 16.8 & 9.2 & 12.2 & 17.1 & 7.3 & 13.3 & 18.9 & 7.7 & 14.1 & 20.3 & 7.9 & 12.5 & 17.4 & 7.5 \\
DecAF~\citep{decaf} & Qwen2.5VL-7B &
23.7 & 30.4 & 17.1 & 19.1 & 24.1 & 14.1 & 19.4 & 25.2 & 13.5 & 21.1 & 27.9 & 14.3 & 17.7 & 22.6 & 12.7 \\
\midrule
\midrule
\rowcolor{myblue} \textbf{ReVEAL} (Ours) & Qwen2.5VL-7B &
\textbf{29.4} & \textbf{39.5} & \textbf{19.2} &
\textbf{22.1} & \textbf{29.5} & \textbf{14.8} &
\textbf{24.0} & \textbf{33.6} & \textbf{14.4} &
\textbf{25.4} & \textbf{35.6} & \textbf{15.3} &
\textbf{22.5} & \textbf{31.6} & \textbf{13.5} \\
\bottomrule
\end{tabular}
}

\par\vspace{2pt}
\setlength{\tabcolsep}{3pt}
\centering
\caption{Main comparison of MLLM-based ReVOS methods.}
\label{tab:main}
\vspace{0.5mm}
\resizebox{\linewidth}{!}{
\begin{tabular}{l l ccc ccc ccc ccc ccc}
\toprule
\multirow{2}{*}{Method} & \multirow{2}{*}{MLLM} &
\multicolumn{3}{c}{Ref-DAVIS} & \multicolumn{3}{c}{ReasonVOS} & \multicolumn{3}{c}{ReVOS {\scriptsize (Overall)}} & \multicolumn{3}{c}{MeViS} & \multicolumn{3}{c}{Ref-YT-VOS} \\
\cmidrule(lr){3-5}\cmidrule(lr){6-8}\cmidrule(lr){9-11}
\cmidrule(lr){12-14}\cmidrule(lr){15-17}
& & $\mathcal{J}$\&$\mathcal{F}$ & $\mathcal{J}$ & $\mathcal{F}$ & $\mathcal{J}$\&$\mathcal{F}$ & $\mathcal{J}$ & $\mathcal{F}$ & $\mathcal{J}$\&$\mathcal{F}$ & $\mathcal{J}$ & $\mathcal{F}$ & $\mathcal{J}$\&$\mathcal{F}$ & $\mathcal{J}$ & $\mathcal{F}$ & $\mathcal{J}$\&$\mathcal{F}$ & $\mathcal{J}$ & $\mathcal{F}$ \\
\midrule
\midrule
\multicolumn{17}{c}{\cellcolor{mygray}\textit{Cross-Model Alignment-Based Methods}} \\
\midrule
LISA~\citep{lai2024lisa} & LLaVA-7B & 64.8 & 62.2 & 67.3 & 31.1 & 29.1 & 33.1 & 40.9 & 39.1 & 42.7 & -- & -- & -- & -- & -- & -- \\
VISA~\citep{yan2024visa} & ChatUniVi-7B & 69.4 & 66.3 & 72.5 & -- & -- & -- & 46.9 & 44.9 & 49.0 & 44.5 & 41.8 & 47.1 & 61.5 & 59.8 & 63.2 \\
VideoLISA~\citep{bai2024videolisa} & LLaVA-Phi-3-V & 68.8 & 64.9 & 72.7 & 47.5 & 45.1 & 49.9 & -- & -- & -- & 44.4 & 41.3 & 47.6 & 63.7 & 61.7 & 65.7 \\
InstructSeg~\citep{instructseg} & Mipha-3B  & 71.1 & 67.3 & 74.9 & -- & -- & -- & 54.5 & 52.0 & 56.9 & -- & -- & -- & 67.5 & 65.4 & 69.5 \\
GLUS~\citep{glus} & LLaVA-7B & -- & -- & -- & 49.9 & 47.5 & 52.4 & 54.9 & 52.4 & 57.3 & 51.3 & 48.5 & 54.2 & 67.3 & 65.5 & 69.0 \\
VRS-HQ~\citep{vrshq} & ChatUniVi-7B & 76.0 & 72.6 & 79.4 & 54.9 & 52.6 & 57.3 & 59.1 & 56.6 & 61.6 & 50.6 & 47.6 & 53.7 & 70.4 & 68.3 & \textbf{72.5} \\ 
RGA3~\citep{rga3} & Qwen2.5VL-7B & 72.8 & 68.3 & 77.3 & 53.6 & 51.3 & 56.0 & 58.0 & 55.9  & 60.0 & 50.1 & 47.4 & 52.8 & 68.5 & 66.8 & 70.1 \\
UniPixel~\citep{unipixel} & Qwen2.5VL-7B & \textbf{76.4} & \textbf{72.7} & \textbf{80.1} & -- & -- & -- & \textbf{63.7} & \textbf{61.7} & \textbf{65.7} & \textbf{55.8} & \textbf{53.2}	 & \textbf{58.3} & \textbf{71.0} & \textbf{69.5}	 & 72.4 \\
Veason-R1~\citep{veasonr1} & Qwen2.5VL-7B & -- & -- & -- & \textbf{59.9} & \textbf{56.0} & \textbf{63.8} & 61.3 & 58.2 & 64.4 & 52.2 & 48.4 & 56.0 & -- & -- & -- \\
\midrule
\multicolumn{17}{c}{\cellcolor{myblue}\textit{Cross-Model Alignment-Free Methods}} \\
\midrule
Loc-Head~\citep{kang2025your} & Qwen2.5VL-7B & 64.6 & 60.2 & 68.9 & 41.1 & 37.9 & 44.3 & 47.0 & 43.3 & 50.7 & 39.4 & 35.2 & 43.6 & 51.0 & 46.8 & 55.2 \\
DecAF~\citep{decaf} & Qwen2.5VL-7B & 75.2 & 70.9 & 79.5 & 63.9 & 60.5 & 67.2 & 54.2 & 50.1 & 58.2 & 48.1 & 44.0 & \textbf{52.1} & 59.9 & 56.2 & 63.5 \\
CoT-RVS~\citep{kao2025cot} & Gemma3-12B & 74.6  & 70.9 &  78.3 & 50.7  & 47.5  & 54.0 & 47.1  & 43.4  & 50.9 & 44.2 & 40.3 & 48.1 & -- & -- & -- \\ 
\midrule
\midrule
\rowcolor{myblue} \textbf{ReVEAL} (Ours) & Qwen2.5VL-7B &
77.7 & 74.1 & 81.3 &
65.1 & 61.9 & 68.3 &
59.0 & 56.3 & 61.6 &
47.7 & 44.7 & 50.6 &
69.2 & 67.2 & 71.1 \\
\rowcolor{myblue} \textbf{ReVEAL} (Ours) & Qwen3VL-8B &
\textbf{80.7} & \textbf{77.3} & \textbf{84.0} &
\textbf{66.1} & \textbf{62.9} & \textbf{69.2} &
\textbf{61.6} & \textbf{58.9} & \textbf{64.3} &
\textbf{49.1} & \textbf{46.2} & 52.0 &
\textbf{71.3} & \textbf{69.3} & \textbf{73.3} \\
\bottomrule
\end{tabular}
}

\end{table*}

\paragraph{Masks with SAM2.}






As shown in Table~\ref{tab:main}, among alignment-free methods, ReVEAL achieves the best performance across all benchmarks. With the same backbone, it outperforms DecAF by 4.8\% and 9.3\%  $\mathcal{J}\&\mathcal{F}$ on ReVOS and Ref-YouTube-VOS, respectively, and consistently outperforms CoT-RVS despite using a smaller backbone. These results suggest that the high-attention regions identified by ReVEAL provide reliable visual prompts for mask generation. 
While alignment-based methods benefit from cross-model optimization between the MLLM and the segmentation module, ReVEAL approaches the performance of RGA-3 under the same Qwen2.5-VL-7B backbone without costly cross-model adapter training. With Qwen3-VL-8B, ReVEAL achieves the highest scores on Ref-DAVIS and ReasonVOS, surpassing the strongest alignment-based results by 4.3\% and 6.2\% $\mathcal{J}\&\mathcal{F}$, respectively, while substantially narrowing the gap on ReVOS, MeViS, and Ref-YouTube-VOS. Additional results are provided in the Appendix~\ref{app:reveal_main}

\subsection{Ablation Study}
\label{sec:experiments:ablation}

In this section, we conduct comprehensive ablation studies. By default, results are reported with ReVEAL-Qwen3VL-8B. Additional results are provided in the Appendix~\ref{app:reveal_ablation}.

\paragraph{Thinking-with-video Paradigm for ReVOS.}

Table~\ref{tab:cot_effect} evaluates the effect of the thinking-with-video paradigm. Without CoT, the model learns to emit VETs within a simple response (e.g., Sure, it is \texttt{<BG>}, \texttt{<NEG>} and \texttt{<OBJ>}.)
With CoT, attention-derived mask $\mathcal{J}\&\mathcal{F}$ improves by 1.1\% and 1.7\% on Ref-DAVIS and ReasonVOS, respectively, while the corresponding SAM2 masks improve by 3.1\% and 12.2\%.
These results show that VET-augmented video reasoning improves grounding beyond simple outputs, providing more effective guidance for downstream segmentation.

\paragraph{Training Components.}
Table~\ref{tab:ablation} shows that SFT alone is insufficient for reliable VET grounding. CAGL-P improves target coverage in latent evidence, while CAGL-R makes attention scores in the fused maps more reliable for point selection. DALR further improves evidence aggregation over fixed-layer selection schemes. Together, these components help ReVEAL elicit faithful role-aware visual evidence during video reasoning to guide segmentation.

\begin{table*}[t!]
\centering
\centering
\begin{minipage}[t]{0.32\linewidth}
\centering
\setlength{\tabcolsep}{4pt}
\caption{Effect of the paradigm.}
\label{tab:cot_effect}
\vspace{0.5mm}
\resizebox{\linewidth}{!}{%
\begin{tabular}{l l cc}
\toprule
\multirow{2}{*}{Mask} & \multirow{2}{*}{\shortstack[c]{Reason\\Type}} &
\multicolumn{2}{c}{$\mathcal{J}\&\mathcal{F}$} \\
\cmidrule(lr){3-4}
& & Ref-D & ReasV \\
\midrule
\midrule
 & w/o CoT & 29.1 & 18.9 \\
 & w/ CoT  & 30.2 & 20.6 \\
\rowcolor{myblue} \multirow{-3}{*}{Attn} & $\Delta$ & \textbf{+1.1} & \textbf{+1.7} \\
\midrule
 & w/o CoT & 77.6 & 53.9 \\
 & w/ CoT  & 80.7 & 66.1 \\
\rowcolor{myblue} \multirow{-3}{*}{SAM2} & $\Delta$ & \textbf{+3.1} & \textbf{+12.2} \\
\bottomrule
\end{tabular}
}

\end{minipage}
\hfill
\begin{minipage}[t]{0.64\linewidth}
\centering
\setlength{\tabcolsep}{4pt}
\caption{Ablation of the training components. First three methods aggregate VET attention maps over six manually selected layers.}
\label{tab:ablation}
\vspace{0.5mm}
\resizebox{\linewidth}{!}{%
\begin{tabular}{ccc cc cc}
\toprule
\multirow{2}{*}{CAGL-P} &
\multirow{2}{*}{CAGL-R} &
\multirow{2}{*}{DALR} &
\multicolumn{2}{c}{Attn Mask} &
\multicolumn{2}{c}{SAM2 Mask} \\
\cmidrule(lr){4-5}\cmidrule(lr){6-7}
& & & Ref-D & ReasV & Ref-D & ReasV \\
\midrule
\midrule
 -- & -- & -- & 9.2 & 6.4 & 32.5 & 22.6 \\
\checkmark & -- & --  & 29.5 & \textbf{22.9} & 76.4 & 61.5 \\
\checkmark & \checkmark & -- & 29.9 & 21.0 & 79.3 & 64.9 \\
\rowcolor{myblue} \checkmark & \checkmark & \checkmark & \textbf{30.2} & 20.6 & \textbf{80.7} & \textbf{66.1} \\
\bottomrule
\end{tabular}
}
\end{minipage}

\par\vspace{3pt}
\setlength{\tabcolsep}{3pt}
\centering
\caption{Comparisons across more MLLM backbones.}
\label{tab:backbone}
\vspace{0.5mm}
\resizebox{\linewidth}{!}{
\begin{tabular}{l l l ccc ccc ccc ccc ccc}
\toprule
\multirow{2}{*}{Method} & \multirow{2}{*}{MLLM} & \multirow{2}{*}{Size} &
\multicolumn{3}{c}{Ref-DAVIS} & \multicolumn{3}{c}{ReasonVOS} & \multicolumn{3}{c}{ReVOS {\scriptsize (Overall)}} & \multicolumn{3}{c}{ReVOS {\scriptsize (Referring)}} & \multicolumn{3}{c}{ReVOS {\scriptsize (Reasoning)}} \\
\cmidrule(lr){4-6}\cmidrule(lr){7-9}\cmidrule(lr){10-12}
\cmidrule(lr){13-15}\cmidrule(lr){16-18}
& & & $\mathcal{J}$\&$\mathcal{F}$ & $\mathcal{J}$ & $\mathcal{F}$ & $\mathcal{J}$\&$\mathcal{F}$ & $\mathcal{J}$ & $\mathcal{F}$ & $\mathcal{J}$\&$\mathcal{F}$ & $\mathcal{J}$ & $\mathcal{F}$ & $\mathcal{J}$\&$\mathcal{F}$ & $\mathcal{J}$ & $\mathcal{F}$ & $\mathcal{J}$\&$\mathcal{F}$ & $\mathcal{J}$ & $\mathcal{F}$ \\
\midrule
\midrule
DecAF & InternVL3 & 8B &
62.8 & 56.9 & 68.6 & 58.9 & 55.1 & 62.7 & 47.4 & 43.7 & 51.2 & 51.7 & 47.9 & 55.5 & 43.2 & 39.5 & 46.8 \\
\rowcolor{myblue} \textbf{ReVEAL} & InternVL3 & 8B &
\textbf{77.8} & \textbf{74.2} & \textbf{81.4} & \textbf{63.6} & \textbf{60.9} & \textbf{66.3} & \textbf{57.4} & \textbf{54.7} & \textbf{60.1} & \textbf{62.1} & \textbf{59.6} & \textbf{64.6} & \textbf{52.7} & \textbf{49.8} & \textbf{55.6} \\
\midrule
DecAF & Qwen2VL & 7B &
64.1 & 59.4 & 68.9 & 52.5 & 49.0 & 56.0 & 45.3 & 41.6 & 49.0 & 52.7 & 48.9 & 56.4 & 37.9 & 34.3 & 41.5 \\
\rowcolor{myblue} \textbf{ReVEAL} & Qwen2VL & 7B &
\textbf{74.9} & \textbf{71.3} & \textbf{78.5} & \textbf{61.2} & \textbf{58.4} & \textbf{64.1} & \textbf{55.9} & \textbf{53.2} & \textbf{58.6} & \textbf{60.7} & \textbf{58.0} & \textbf{63.3} & \textbf{51.1} & \textbf{48.4} & \textbf{53.8} \\
\midrule
DecAF & Qwen2.5VL & 3B &
58.1 & 53.6 & 62.6 & 52.9 & 49.6 & 56.2 & 39.7 & 35.6 & 43.8 & 46.5 & 42.1 & 50.9 & 32.9 & 29.1 & 36.7 \\
\rowcolor{myblue} \textbf{ReVEAL} & Qwen2.5VL & 3B &
\textbf{74.5} & \textbf{70.2} & \textbf{78.8} & \textbf{54.0} & \textbf{51.3} & \textbf{56.8} & \textbf{51.6} & \textbf{48.6} & \textbf{54.7} & \textbf{56.9} & \textbf{54.1} & \textbf{59.6} & \textbf{46.4} & \textbf{43.0} & \textbf{49.7} \\
\midrule
\rowcolor{myblue} \textbf{ReVEAL} & Qwen3VL & 4B &
77.1 & 73.4 & 80.8 & 60.5 & 57.6 & 63.4 & 58.0 & 55.3 & 60.8 & 62.2 & 59.7 & 64.7 & 53.9 & 50.9 & 56.9
\\
\rowcolor{myblue} \textbf{ReVEAL} & Qwen3VL & 8B &
\textbf{80.7} & \textbf{77.3} & \textbf{84.0} & \textbf{66.1} & \textbf{62.9} & \textbf{69.2} & \textbf{61.6} & \textbf{58.9} & \textbf{64.3} & \textbf{65.0} & \textbf{62.5} & \textbf{67.5} & \textbf{58.3} & \textbf{55.4} & \textbf{61.1} \\
\bottomrule
\end{tabular}
}
\par\vspace{5pt}
\centering
\begin{minipage}[t]{0.38\linewidth}
\centering
\setlength{\tabcolsep}{4pt}
\caption{Ablation across different segmentation backbones.}
\label{tab:seg_model}
\vspace{0.5mm}
\resizebox{\linewidth}{!}{
\begin{tabular}{l ccc}
\toprule
Model & Ref-D & ReasV & Ref-YT \\
\midrule
\midrule
SAM2-Hiera-T & 79.1 & 64.0 & 67.5 \\
SAM2-Hiera-S & 78.7 & 64.3 & 69.0 \\
SAM2-Hiera-B  & 79.9 & 67.0 & 70.3 \\
SAM2-Hiera-L & \textbf{80.7} & 66.1 & 71.3 \\
\rowcolor{myblue} \textbf{SAM3}  & 79.6 & \textbf{67.6} & \textbf{71.4} \\
\bottomrule
\end{tabular}
}

\end{minipage}
\hfill
\begin{minipage}[t]{0.59\linewidth}
\centering
\setlength{\tabcolsep}{4pt}
\caption{Inference-time ablation of different VET attention map fusion strategies. Here, we report the $\mathcal{J}\&\mathcal{F}$ metric.}
\label{tab:tokencomb}
\vspace{0.5mm}
\resizebox{\linewidth}{!}{
\begin{tabular}{l ccc cc cc}
\toprule
\multirow{2}{*}{MLLM} &
\multirow{2}{*}{\texttt{<OBJ>}} &
\multirow{2}{*}{\texttt{<BG>}} &
\multirow{2}{*}{\texttt{<NEG>}} &
\multicolumn{2}{c}{Attn Mask} &
\multicolumn{2}{c}{SAM2 Mask} \\
\cmidrule(lr){5-6}\cmidrule(lr){7-8}
& & & & Ref-D & ReasV & Ref-D & ReasV \\
\midrule
\midrule
\multirow{4}{*}{Qwen3VL-8B} & \checkmark & -- & -- & 27.2 & 18.8 & 78.4 & 63.2 \\
& \checkmark & -- & \checkmark & 28.8 & 19.6 & \textbf{80.7} & 65.8 \\
& \checkmark & \checkmark & -- & \textbf{31.1} & \textbf{23.4} & 78.5 & 64.4 \\
\rowcolor{myblue} & \checkmark & \checkmark & \checkmark & 30.2 & 20.6 & \textbf{80.7} & \textbf{66.1} \\
\bottomrule
\end{tabular}
}
\end{minipage}

\end{table*}

\paragraph{VET Fusion at Inference.}


Table~\ref{tab:tokencomb} compares different VET attention combinations from the same CoT at inference. Adding \texttt{<BG>} improves attention-mask scores by 3.9\% and 4.6\% on Ref-DAVIS and ReasonVOS, while adding \texttt{<NEG>} instead yields better SAM2 segmentation despite lower attention-mask scores. This suggests that background suppression refines the overall target coverage, whereas distractor suppression makes attention peaks more discriminative. 
Combining all three VET types yields the best overall performance, supporting the effectiveness of our role design.

\paragraph{MLLM Backbone Scalability.}

As shown in Table~\ref{tab:backbone}, ReVEAL consistently outperforms the baseline across diverse MLLM backbones, demonstrating its generalizability.
We further observe that the performance generally improves with backbone size. For example, scaling Qwen3VL from 4B to 8B improves  $\mathcal{J}\&\mathcal{F}$ by 3.6\% on Ref-DAVIS and ReVOS and 5.6\% on ReasonVOS, suggesting that larger MLLMs provide more reliable visual evidence for downstream segmentation.

\paragraph{Segmentation Backbone Scalability.}

ReVEAL uses latent visual evidence from MLLM reasoning to guide segmentation without requiring alignment to a specific segmenter. As shown in Table~\ref{tab:seg_model}, the latent grounding evidence generalizes across smaller SAM2 variants and SAM3~\citep{sam3} on all three benchmarks, with minor performance variations due to the segmenter's own capability.

\begin{figure*}[t]
  \centering
  \includegraphics[width=\textwidth]{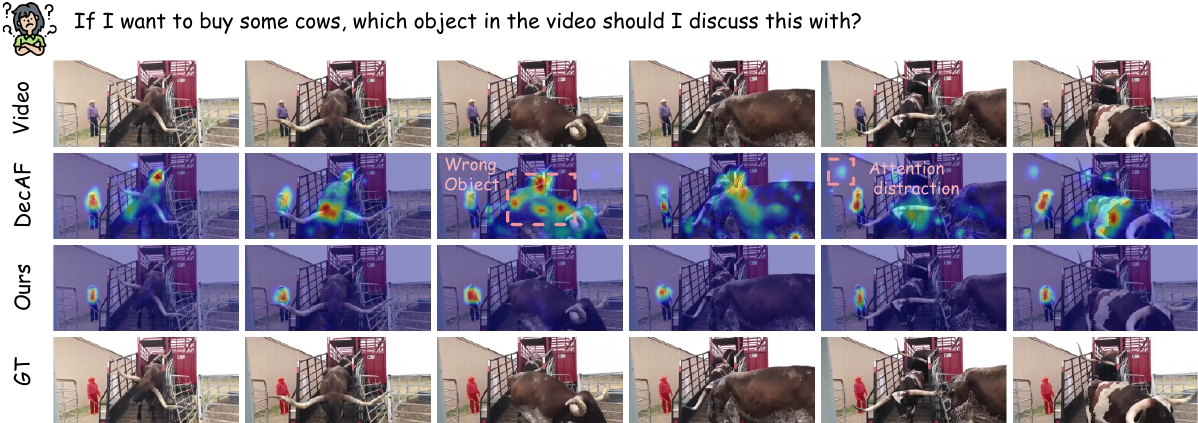}
  \setlength{\abovecaptionskip}{0pt}
  \vspace{-3pt}
  \caption{Qualitative comparison of ReVEAL and DecAF on a query requiring reasoning.}
  \label{fig:qualitative}
\end{figure*}

\begin{figure*}[t]
  \centering
  \includegraphics[width=\textwidth]{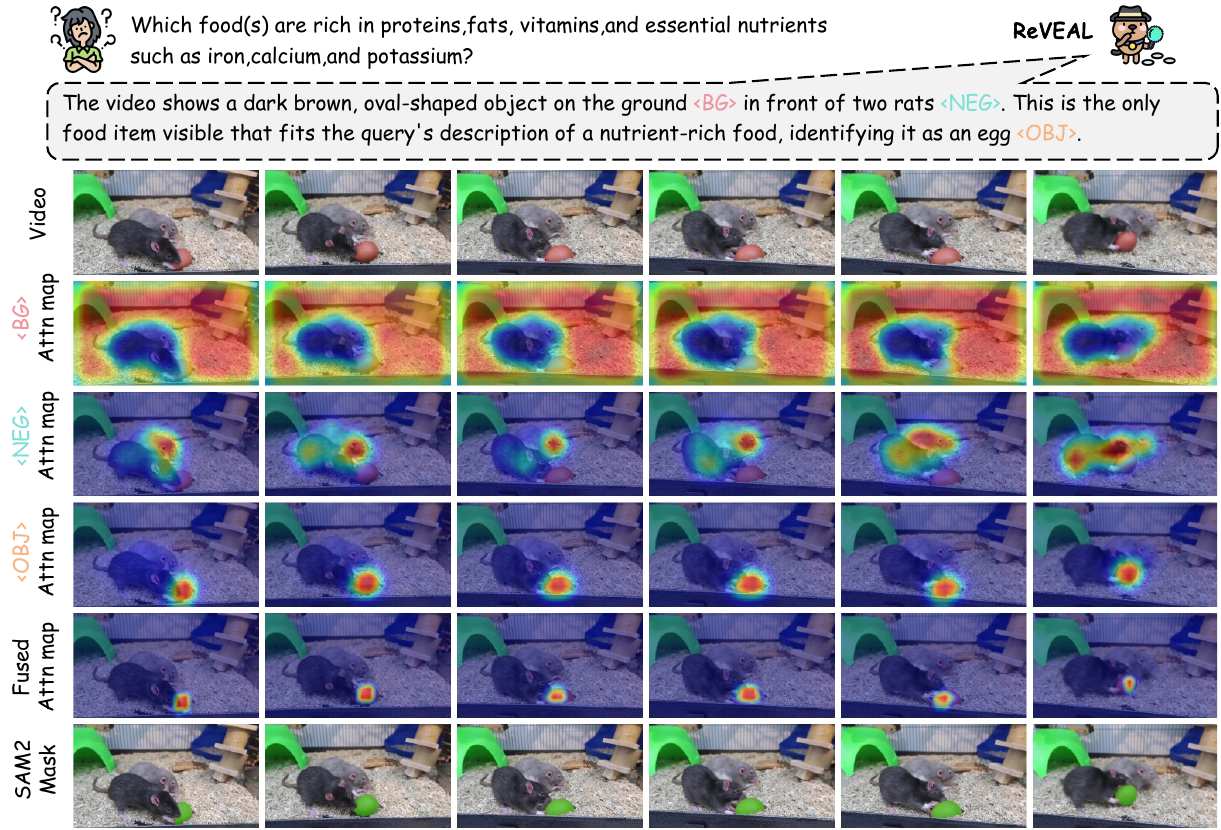}
  \setlength{\abovecaptionskip}{0pt}
  \vspace{-3pt}
  \caption{Visualization of ReVEAL's response and role-specific VET attention maps.}
  \label{fig:token_attn}

\end{figure*}

\paragraph{Qualitative Analysis.}
We provide qualitative analysis of ReVEAL from two perspectives. First, Fig.~\ref{fig:qualitative} shows that ReVEAL reduces noise in the fused attention maps compared with DecAF, resulting in more precise focus on the target object and more accurate grounding for challenging queries requiring implicit reasoning.
%
Second, Fig.~\ref{fig:token_attn} shows that different VETs generally attend to regions consistent with their intended roles.
Additional cases are provided in Appendix~\ref{app:visualizations}.

\section{Conclusion}


We presented ReVEAL, a decoupled thinking-with-video framework for alignment-free ReVOS. ReVEAL enables the MLLM to identify targets through video reasoning and derive grounding cues from role-aware VET attention, without requiring cross-model alignment with the segmenter during training. At inference, ReVEAL derives role-specific evidence from VET-augmented video reasoning to guide a frozen SAM2 for target mask prediction. Experiments across five RVOS and ReVOS benchmarks demonstrate the effectiveness and generalizability of ReVEAL. Overall, our work highlights the potential of internal signals from MLLM reasoning to bridge high-level video understanding and dense visual prediction.
\section*{AI Use Statement}
\label{sec:llm}

We use generative AI tools in constructing the ReVEAL-47K training dataset, including video CoT annotation and formatting collected samples into VET-augmented CoT traces, as described in Section~\ref{sec:method:approach} and the Appendix. GPT is used to polish the manuscript, including language refinement and grammar checking of our original draft. We do not use AI tools for other research tasks requiring disclosure. We review all AI-assisted content and take full responsibility for the final paper.

\bibliographystyle{preprint}
\bibliography{bibliography}

\appendix
\newpage
\appendix

\section*{Appendix Overview}

This appendix provides supplementary details of ReVEAL.

\begingroup
\hypersetup{hidelinks}
\makeatletter
\renewcommand*{\l@section}[2]{%
  \addvspace{1em}%
  \begingroup
    \normalfont\large\scshape
    \@dottedtocline{1}{0pt}{1.5em}{#1}{#2}%
  \endgroup
}
\renewcommand*{\l@subsection}[2]{%
  \begingroup
    \normalfont\normalsize\scshape
    \@dottedtocline{2}{1.5em}{2.3em}{#1}{#2}%
  \endgroup
}
\makeatother
\newcommand{\overviewentry}[3]{%
  \csname l@#1\endcsname
    {\hyperref[#2]{\numberline{\ref*{#2}}#3}}%
    {\hyperref[#2]{\pageref*{#2}}}%
}
\overviewentry{section}{app:dataset}{ReVEAL-47K Training Dataset}
\overviewentry{subsection}{app:dataset:overview}{Dataset Overview}
\overviewentry{subsection}{app:dataset:prompt}{Annotation Prompts}

\overviewentry{section}{app:details}{More Details of ReVEAL}

\overviewentry{subsection}{app:router}{Dynamic Attention Layer Router}
\overviewentry{subsection}{app:prompt}{ReVEAL System Prompt}
\overviewentry{subsection}{app:training}{Training and Inference Settings}

\overviewentry{section}{app:more_results}{Additional Experimental Analysis}
\overviewentry{subsection}{app:reveal_main}{Additional Quantitative Results}
\overviewentry{subsection}{app:reveal_ablation}{Additional Ablation Studies}
\overviewentry{section}{app:visualizations}{More Visualizations}
\overviewentry{subsection}{app:vis:decaf}{More Comparisons with Other Methods}
\overviewentry{subsection}{app:vis:roles}{More Role-Token Attention Map Visualizations}
\overviewentry{subsection}{app:vis:failure}{Failure Case Analysis and Limitations}
\endgroup
\clearpage

\section{ReVEAL-47K Training Dataset}
\label{app:dataset}

\subsection{Dataset Overview}
\label{app:dataset:overview}


\begin{wrapfigure}{r}{0.42\textwidth}
  \centering
  \vspace{-\intextsep}
  \includegraphics[width=0.42\textwidth]{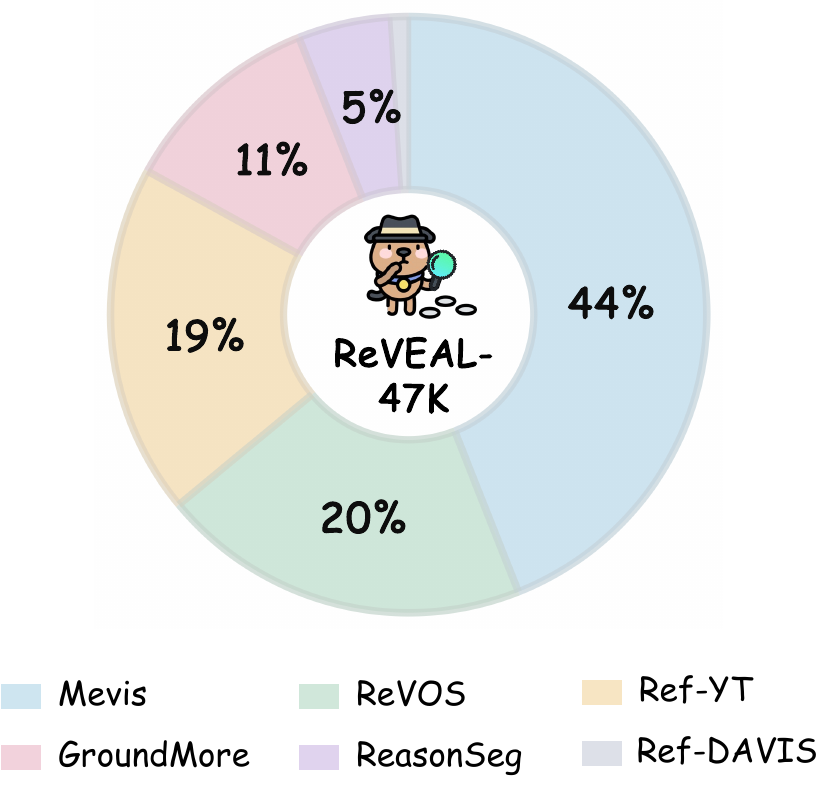}
  \setlength{\abovecaptionskip}{0pt}
  \vspace{-5pt}
  \caption{ReVEAL-47K Composition.}
  \label{fig:sft_data}
\end{wrapfigure}

To equip the model with dynamic role-aware VET reasoning and reliable latent visual grounding, we construct ReVEAL-47K training dataset, comprising 47K VET-augmented ReVOS CoT traces paired with target and background masks for attention grounding.
%

ReVEAL-47K is constructed from the training splits of five public RVOS and ReVOS datasets, including MeViS~\citep{ding2023mevis}, ReVOS~\citep{yan2024visa}, GroundMoRe~\citep{groundmore}, Ref-DAVIS~\citep{davis}, and Ref-YouTube-VOS~\citep{ytvos}, together with the image reasoning segmentation dataset ReasonSeg~\citep{lai2024lisa}, as shown in Fig.~\ref{fig:sft_data}. We use six frames as the model input for all samples. For ReasonSeg, each sample is included twice, with its single image repeated across six frames and treated as a static video. 
The resulting dataset contains 47K VET-augmented CoT annotations. Each trace contains exactly one \texttt{<OBJ>} token for target evidence, even when the query refers to multiple target objects, as its attention map can cover multiple target regions, while suppressive VETs are introduced only when needed. Specifically, \texttt{<BG>} appears at most once to provide background evidence, whereas multiple \texttt{<NEG>} tokens can be used to distinguish different distractors. On average, each trace contains 1.00 \texttt{<OBJ>}, 0.42 \texttt{<BG>}, and 0.86 \texttt{<NEG>} tokens.

\subsection{Annotation Prompts}
\label{app:dataset:prompt}

\paragraph{Stage 1: Mask-conditioned Target Description.}
We use GAR~\citep{wang2025grasp} with the official inference setup to generate a mask-conditioned caption of the target object(s).

\paragraph{Stage 2: Original Video CoT Generation.}
This prompt is used for Gemini2.5 Pro~\citep{comanici2025gemini} to produce a naive CoT for the ReVOS query based on the video content:

\begin{tcolorbox}[promptbox,title={Original Video CoT Generation Prompt}]
\footnotesize\raggedright
You are annotating video object question answering data.\\[3pt]

\textbf{Inputs}
\begin{enumerate}[leftmargin=1.4em,itemsep=1pt,topsep=2pt]
  \item A set of video frames.
  \item A referring query.
  \item A detailed description of the target object(s).
\end{enumerate}

\textbf{Task}\\
Write one concise reasoning paragraph explaining how the target described
by the query can be identified and localized based on visual evidence in the
video. Focus on observable cues such as appearance, position, movement,
persistence across frames, and spatial relations to visible scene elements.
If the target is trivially identifiable because it is unique, salient, and has
no competing objects, one sentence naming the target is enough. Otherwise,
include only the minimum cues needed to remove ambiguity, choosing from:
background regions to ignore, concrete distractors that could be confused with
the target, visible reference objects, temporal order, motion trajectory,
persistence, or key-frame changes. Use only cues that are actually visible and
helpful for this case. Do not pad, do not list absent elements, and do not
invent distractors or references.\\[3pt]

\textbf{Constraints}
\begin{itemize}[leftmargin=1.4em,itemsep=1pt,topsep=2pt]
  \item The target must follow the referring query exactly; do not re-interpret
  or substitute it.
  \item Keep the paragraph concise: 2--5 sentences, information-dense.
  \item Output plain text only. No JSON, no coordinates, no bounding boxes.
\end{itemize}
\end{tcolorbox}

\paragraph{Stage 3: VET-Augmented Reasoning Trace Construction.}

Given the naive CoT traces, we prompt Gemini2.5 Pro~\citep{comanici2025gemini} to refine the reasoning by identifying key objects, background, and distractors, and inserting the corresponding role-specific VETs into the trace:

\begin{tcolorbox}[promptbox,title={VET-Augmented Reasoning Trace Construction Prompt}]
\footnotesize\raggedright
You are rewriting reasoning traces for video object discovery annotations.\\[3pt]

\textbf{Inputs}
\begin{enumerate}[leftmargin=1.4em,itemsep=1pt,topsep=2pt]
  \item A reasoning query.
  \item An existing reasoning paragraph of the video.
\end{enumerate}

\textbf{Task}\\
Rewrite the reasoning paragraph into one concise, natural visual reasoning trace
that answers the reasoning query. Use the existing reasoning paragraph as
evidence, but make the rewritten output focus on identifying and localizing the
object(s) asked for by the query in the video. Do not merely polish the original
text or insert tokens mechanically. The reasoning should proceed from visible
video cues to the final answer: first explain which visual evidence supports the
queried object(s), then conclude with the target object(s) marked by \texttt{<OBJ>}.\\[3pt]

\textbf{Visual Evidence Tokens}\\
Insert Visual Evidence Tokens immediately after the phrase they label:
\begin{itemize}[leftmargin=1.4em,itemsep=1pt,topsep=2pt]
  \item \texttt{<BG>}: background scene regions to ignore, not foreground objects.
  \item \texttt{<NEG>}: concrete non-target foreground objects that could be
  confused with the target.
  \item \texttt{<OBJ>}: the final target object(s) to segment.
\end{itemize}

\textbf{Reasoning Order}\\
Prefer this order when applicable:
\begin{enumerate}[leftmargin=1.4em,itemsep=1pt,topsep=2pt]
  \item Interpret the query by focusing on the object(s) it asks to segment.
  \item Use observable video cues to identify and localize that object(s), such as
  appearance, position, persistence across frames, motion, relation to stable
  scene anchors, or temporal changes.
  \item Add \texttt{<BG>} only for background scene regions that help exclude
  irrelevant areas.
  \item Add \texttt{<NEG>} only for visible non-target foreground objects that
  could plausibly be confused with the queried object.
  \item End with the target that answers the query, marked once with
  \texttt{<OBJ>}.
\end{enumerate}
The paragraph should provide a video-grounded response to the query, rather than a static image description or a list of labels. \\[3pt]

\textbf{Token Placement Rule}\\
Each Visual Evidence Token must appear immediately after the object or region phrase it
labels. Correct: the driveway \texttt{<BG>}, the soldier \texttt{<NEG>}, the man
on the left \texttt{<OBJ>}.\\[3pt]

\textbf{Editing Rules}
\begin{itemize}[leftmargin=1.4em,itemsep=1pt,topsep=2pt]
  \item The rewritten paragraph must answer the reasoning query, not just
  restate the existing paragraph.
  \item The object(s) marked with \texttt{<OBJ>} must be the final answer to the
  query.
  \item Any reasoning before \texttt{<OBJ>} should explain why that object(s),
  rather than another visible object(s) or region, satisfies the query.
  \item Do not mark an object(s) with \texttt{<OBJ>} only because it is visually
  salient; it must be the object(s) asked for by the query.
  \item Do not add new visual objects, distractors, references, background regions, actions, or attributes unless they are already stated or clearly
  implied by the reasoning paragraph.
  \item If the existing paragraph contains irrelevant details that do not help
  answer the query, omit them.
  \item Improve the reasoning so it sounds like it is derived from video
  evidence, not like a labeling checklist.
  \item Do not use meta-descriptions such as ``initial context is established,''
  ``the label matches,'' or ``this indicates the target.''
  \item \texttt{<OBJ>} is mandatory and must appear exactly once, attached to the
  final target.
  \item \texttt{<BG>} is optional and may appear at most once. \texttt{<NEG>} is
  optional and may appear multiple times, once for each distinct distractor.
  Use them only when useful and visually grounded.
  \item Use \texttt{<BG>} only for scene or background regions, never foreground
  objects.
  \item Use \texttt{<NEG>} only for concrete non-target foreground objects that
  could be mistaken for the target.
  \item Do not attach tokens to verbs, actions, relations, attributes, or
  abstract descriptions.
  \item Keep the output concise, fluent, and grounded in the video.
  \item Output only the rewritten reasoning paragraph. No JSON, no bullets, no
  headings, no explanation.
\end{itemize}
\end{tcolorbox}

\paragraph{Stage 4: Quality control.}
Here is the prompt we use to filter invalid CoT annotations:

\begin{tcolorbox}[promptbox,title={Quality Control Prompt}]
\footnotesize\raggedright
You are validating video object reasoning annotations.\\[3pt]

\textbf{Inputs}
\begin{enumerate}[leftmargin=1.4em,itemsep=1pt,topsep=2pt]
  \item Six uniformly sampled video frames.
  \item A referring query.
  \item A target description.
  \item A candidate reasoning trace containing Visual Evidence Tokens.
\end{enumerate}

\textbf{Visual Evidence Tokens}
\begin{itemize}[leftmargin=1.4em,itemsep=1pt,topsep=2pt]
  \item \texttt{<OBJ>}: the final target object.
  \item \texttt{<NEG>}: visible distractors or non-target objects.
  \item \texttt{<BG>}: safe background or regions to ignore.
\end{itemize}

\textbf{Task}\\
Determine whether the reasoning trace correctly identifies and localizes the target object(s). Use the referring query as the primary criterion and the target description only as supplementary guidance. Base the judgment solely on the visible frames.\\[3pt]

\textbf{Answer yes only if}
\begin{itemize}[leftmargin=1.4em,itemsep=1pt,topsep=2pt]
  \item the target (s) is visible in the six frames;
  \item the trace follows the query and matches the target description;
  \item the trace marks the correct final target (s);
  \item any \texttt{<BG>} or \texttt{<NEG>} Visual Evidence Tokens are visually grounded and
  correctly used;
  \item the reasoning is clear, specific, and free of substantial irrelevant or
  redundant content.
\end{itemize}

\textbf{Answer no if}
\begin{itemize}[leftmargin=1.4em,itemsep=1pt,topsep=2pt]
  \item the target (s) is not visible in the six frames;
  \item the trace targets the wrong object or conflicts with the query;
  \item the trace conflicts with the target description;
  \item the trace contradicts the frames or invents key evidence;
  \item the trace is too ambiguous to localize the target;
  \item the trace contains substantial irrelevant, redundant, or generic content.
\end{itemize}

\textbf{Constraints}\\
Output exactly one lowercase word: yes or no. 
\end{tcolorbox}

\section{More Details of ReVEAL}
\label{app:details}

\subsection{Dynamic Attention Layer Router}
\label{app:router}


Since target identification is central to ReVOS, DALR uses \texttt{<OBJ>} attention as the routing cue to select layers that provide informative target evidence. As these layers vary across inputs, DALR adaptively selects and weights candidate layers for each sample. The same layer selection and routing weights are applied to all VET roles, ensuring efficient aggregation of role-specific attention maps.
As detailed in Algorithm~\ref{alg:router}, DALR employs noisy top-$k$ routing during training to encourage exploration and deterministic routing at inference. We use $L_c=14$ candidate layers for Qwen2.5VL-7B and $L_c=18$ for Qwen3VL-8B, with a feature dimension of $256$ and $k=6$ by default.

\begin{algorithm}[H]
\caption{Dynamic Attention Layer Router}
\label{alg:router}
\begin{algorithmic}[1]
\Require \texttt{<OBJ>} attention maps
$a^{\texttt{<OBJ>},1}$ from candidate layers $\mathcal{L}_c$,
feature dimension $D$, and selection size $k$
\State $\mathbf{X}\gets
\operatorname{ClampMin}
\bigl(\operatorname{Flatten}(a^{\texttt{<OBJ>},1}),0\bigr)$
\State $\mathbf{X}\gets\operatorname{Normalize}(\mathbf{X})$
\State $\mathbf{F}\gets
\operatorname{AdaptiveAvgPool1D}(\mathbf{X},D)$
\State $\mathbf{F}\gets\operatorname{Normalize}(\mathbf{F})$
\State $\mathbf{F}\gets\operatorname{StopGradient}(\mathbf{F})$
\State $\mathbf{s}\gets g_{\mathrm{score}}(\mathbf{F})$
\If{training}
    \State $\boldsymbol{\sigma}\gets
    \operatorname{Softplus}\bigl(g_{\mathrm{noise}}(\mathbf{F})\bigr)$
    \State $\tilde{\mathbf{s}}\gets
    \mathbf{s}+\boldsymbol{\epsilon}\odot\boldsymbol{\sigma},
    \quad
    \boldsymbol{\epsilon}\sim\mathcal{N}(\mathbf{0},\mathbf{I})$
\Else
    \State $\tilde{\mathbf{s}}\gets\mathbf{s}$
\EndIf
\State $\mathcal{S}\gets
\operatorname{TopK}_{l\in\mathcal{L}_c}(\tilde{s}_l,k)$
\State $w_l\gets
\displaystyle\frac{\exp(\tilde{s}_l)}
{\sum_{m\in\mathcal{S}}\exp(\tilde{s}_m)},
\quad \forall l\in\mathcal{S}$
\State \Return selected layers $\mathcal{S}$ and routing weights
$\{w_l\}_{l\in\mathcal{S}}$
\end{algorithmic}
\end{algorithm}

\subsection{ReVEAL System Prompt}
\label{app:prompt}

We use the following system prompt during both training and inference:

\begin{tcolorbox}[promptbox,title={ReVEAL System Prompt}]
\footnotesize\raggedright
You are a video object reasoner.\\[3pt]

\textbf{Task}\\
Given a set of video frames and a referring query, your job is to answer the
query by locating the target object(s) in the video. Output a reasoning paragraph
that answers the query and contains the following three special tokens.\\[3pt]

\textbf{Visual Evidence Tokens}
\begin{itemize}[leftmargin=1.4em,itemsep=1pt,topsep=2pt]
  \item \texttt{<BG>}: background regions to ignore, not foreground objects.
  \item \texttt{<NEG>}: distractor objects that look similar to the target.
  \item \texttt{<OBJ>}: the final target to segment.
\end{itemize}

\textbf{Output}\\
\texttt{<BG>} is optional and may appear at most once. \texttt{<NEG>} is optional
and may appear multiple times, once for each distinct distractor; these tokens
may appear in any order. End with a sentence that directly answers the query and names the target with
\texttt{<OBJ>}. Output only the paragraph: no lists, coordinates, masks, or JSON.
\end{tcolorbox}

\subsection{Training and Inference Settings}
\label{app:training}

Table~\ref{tab:train_hparams} and ~\ref{tab:infer_hparams} provide key hyperparameters in training and evaluating ReVEAL-Qwen3VL-8B.

\begin{table}[h]
\centering
\begin{minipage}[t]{0.5\linewidth}
\centering
\caption{Training settings.}
\label{tab:train_hparams}
\vspace{0.5mm}
\begin{tabular}{l |l}
\toprule
Hyperparameter & Value \\
\midrule
Random Seed & 42 \\
Trainable Modules & MLLM, DALR \\
Optimizer & AdamW \\
Epoch & 1 \\
Learning Rate & $2\times10^{-5}$ \\
Weight Decay & $0.05$ \\
LR Schedule & Cosine \\
Global Batch Size & 16 \\
Frame Max Pixels & $336 \times 32 \times 32$  \\
Precision & bfloat16 \\
DALR Top-$k$ & 6 \\
CAGL Weight $\lambda_{\mathrm{CAGL}}$  & 0.2 \\
CAGL-R Weight $\lambda_{\mathrm{rep}}$ & 0.1 \\
\bottomrule 
\end{tabular}
\end{minipage}
\hfill
\begin{minipage}[t]{0.47\linewidth}
\centering
\caption{Inference settings.}
\label{tab:infer_hparams}
\vspace{0.5mm}
\begin{tabular}{l |l}
\toprule
Hyperparameter & Value \\
\midrule
\texttt{<BG>} Sup Weight $\alpha_{\texttt{<BG>}}$  & 0.1 \\
\texttt{<NEG>} Sup Weight $\alpha_{\texttt{<NEG>}}$ & 1.0 \\
Max Point Prompts & 3 \\
Attention Threshold $\tau_{\mathrm{pq}}$ & 0.8 \\
NMS IoU Threshold $\tau_{\mathrm{nms}}$ & 0.7 \\
Tracklet-consistency Threshold $\tau_{\mathrm{trk}}$ & 0.8 \\
\bottomrule 
\end{tabular}
\par\vspace{15pt}
\centering
\setlength{\abovecaptionskip}{0pt}
\caption{
Per-query MLLM inference time on Ref-DAVIS using the same Qwen2.5-VL-7B.
}
\label{tab:speed}
\vspace{0.5mm}

\resizebox{\linewidth}{!}{
\begin{tabular}{lcc}
\toprule
Method & MLLM calls & Avg. time (s) \\
\midrule
DecAF & 35 & 23.5 \\
\rowcolor{myblue}
\textbf{ReVEAL} (Ours) & \textbf{1} & \textbf{5.8~($4.1\times$)} \\
\bottomrule
\end{tabular}
}

\end{minipage}
\end{table}

\section{Additional Experimental Analysis}
\label{app:more_results}

By default, results are reported with ReVEAL-Qwen3VL-8B and the
$\mathcal{J}\&\mathcal{F}$ metric.

\subsection{Additional Quantitative Results}
\label{app:reveal_main}

\paragraph{Inference Efficiency.}
We compare the efficiency of MLLM-side attention production of ReVEAL with the baseline DecAF~\citep{decaf} on Ref-DAVIS using the same Qwen2.5VL-7B. As shown in Table~\ref{tab:speed}, ReVEAL requires only one MLLM response per query, compared with 35 for DecAF, and achieves over \(4\times\) faster inference. 
Measurement is conducted on an NVIDIA H800 GPU.

\paragraph{Motion-grounded Video Reasoning Performance.}


GroundMoRe~\citep{groundmore} is a challenging benchmark that requires joint spatio-temporal reasoning to understand object motions and interactions from causal, sequential, counterfactual, and descriptive perspectives. The results in Table~\ref{tab:groundmore} show that ReVEAL achieves an overall $\mathcal{J}\&\mathcal{F}$ result of 37.1\%, outperforming MORA (27.2\%) by 9.9\%. Notably, ReVEAL delivers substantial gains on both Causal and Descriptive questions, improving the scores from 27.4\% to 37.7\% and from 29.6\% to 44.8\%, respectively.
These gains suggest that our thinking-with-video paradigm facilitates causal reasoning, while attention-derived latent evidence provides a video-wide perspective for tracking moving targets more accurately.

\begin{table*}[t!]
\setlength{\tabcolsep}{3pt}
\centering
\caption{Comparison on GroundMoRe.}
\label{tab:groundmore}
\vspace{0.5mm}
\resizebox{\linewidth}{!}{
\begin{tabular}{l ccc ccc ccc ccc ccc}
\toprule
\multirow{2}{*}{Method} &
\multicolumn{3}{c}{Overall} &
\multicolumn{3}{c}{Causal} &
\multicolumn{3}{c}{Sequential} &
\multicolumn{3}{c}{Counterfactual} &
\multicolumn{3}{c}{Descriptive} \\
\cmidrule(lr){2-4}\cmidrule(lr){5-7}\cmidrule(lr){8-10}
\cmidrule(lr){11-13}\cmidrule(lr){14-16}
& $\mathcal{J}$\&$\mathcal{F}$ & $\mathcal{J}$ & $\mathcal{F}$ &
$\mathcal{J}$\&$\mathcal{F}$ & $\mathcal{J}$ & $\mathcal{F}$ &
$\mathcal{J}$\&$\mathcal{F}$ & $\mathcal{J}$ & $\mathcal{F}$ &
$\mathcal{J}$\&$\mathcal{F}$ & $\mathcal{J}$ & $\mathcal{F}$ &
$\mathcal{J}$\&$\mathcal{F}$ & $\mathcal{J}$ & $\mathcal{F}$ \\
\midrule

\multicolumn{16}{c}{\cellcolor{mygray}\textit{None-MLLM RVOS Methods}} \\
\midrule
ReferFormer~\citep{wu2022referformer} &
12.7 & 11.2 & 14.3 &
12.7 & 11.0 & 14.4 &
10.7 & 9.4 & 12.1 &
12.4 & 11.1 & 13.6 &
15.2 & 13.3 & 17.0 \\

SgMg~\citep{sgmg} &
17.5 & 15.8 & 19.2 &
18.9 & 17.0 & 20.9 &
15.5 & 14.0 & 17.0 &
16.4 & 15.2 & 17.5 &
18.8 & 16.7 & 20.9 \\

HTR~\citep{htr} &
14.5 & 12.7 & 16.2 &
15.7 & 13.8 & 17.5 &
12.5 & 10.8 & 14.1 &
13.5 & 12.2 & 14.9 &
15.9 & 13.8 & 18.1 \\

LMPM~\citep{ding2023mevis} &
13.3 & 12.7 & 14.0 &
13.7 & 13.1 & 14.3 &
11.3 & 10.6 & 12.0 &
13.2 & 12.6 & 13.7 &
14.3 & 13.6 & 15.1 \\

\midrule
\multicolumn{16}{c}{\cellcolor{mygray}\textit{Image Reasoning Segmentation Methods}} \\
\midrule

LISA~\citep{lai2024lisa} &
6.1 & 4.9 & 7.3 &
5.5 & 4.4 & 6.7 &
5.2 & 4.1 & 6.3 &
5.9 & 4.8 & 7.0 &
7.6 & 6.1 & 9.0 \\

PixelLM-7B~\citep{pixellm} &
10.0 & 9.9 & 10.0 &
9.5 & 9.4 & 9.7 &
9.2 & 9.2 & 9.3 &
9.8 & 9.8 & 9.9 &
10.9 & 11.0 & 10.9 \\

\midrule
\multicolumn{16}{c}{\cellcolor{mygray}\textit{MLLM-Based ReVOS Methods}} \\
\midrule


VISA-7B~\citep{yan2024visa} &
5.3 & 4.7 & 5.9 &
6.1 & 5.5 & 6.6 &
4.5 & 3.8 & 5.2 &
5.1 & 4.3 & 5.9 &
5.4 & 5.0 & 5.8 \\

MORA~\citep{groundmore} &
27.2 &  27.4 & 26.9 &
27.4 & 26.9 & 28.0 &
26.1  & 27.2  & 24.9 &
25.6  & 25.3 &  25.9 &
29.6  & 30.2 &  28.9 \\

\midrule
\midrule


\rowcolor{myblue}
\textbf{ReVEAL} (Ours) &
\textbf{37.1} & \textbf{35.8} & \textbf{38.4} &
\textbf{37.7} & \textbf{36.4} & \textbf{39.1} &
\textbf{34.0} & \textbf{32.7} & \textbf{35.3} &
\textbf{31.5} & \textbf{30.3} & \textbf{32.7} &
\textbf{44.8} & \textbf{43.3} & \textbf{46.2} \\
					
\bottomrule
\end{tabular}
}
\end{table*}

\subsection{Additional Ablation Studies}
\label{app:reveal_ablation}

\paragraph{Sensitivity to the CAGL Weights.}
Table~\ref{tab:cagl_weight} evaluates the sensitivity of CAGL to its loss weights. Overall, ReVEAL maintains relatively stable performance across different weight configurations, indicating robustness to these hyperparameters. We adopt $\lambda_{\mathrm{CAGL}}=0.2$ and $\lambda_{\mathrm{rep}}=0.1$, which achieves the best performance on both benchmarks.

\begin{figure}[ht]
  \centering
  \includegraphics[width=0.6\linewidth]{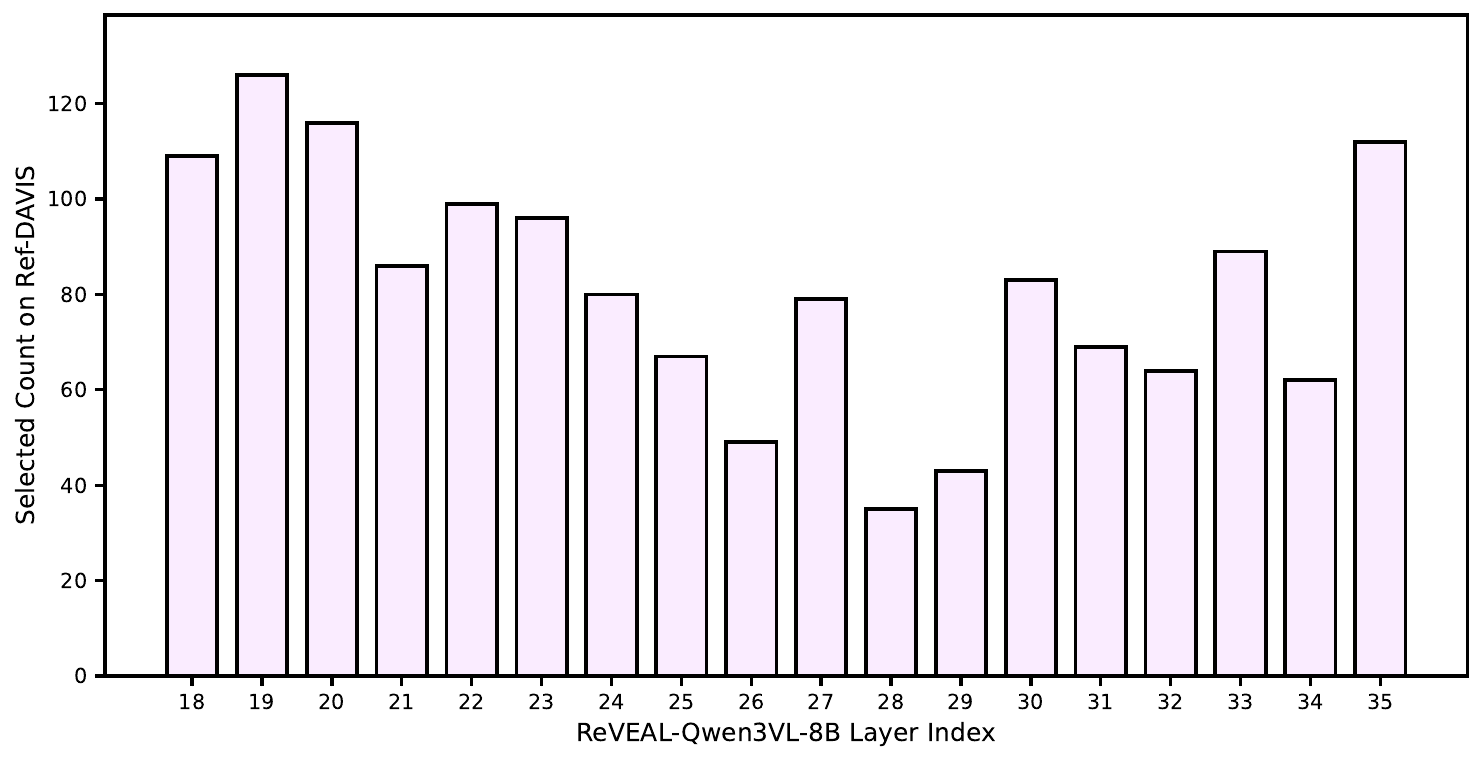}
  \setlength{\abovecaptionskip}{0pt}
  \caption{Layer selection distribution of DALR in ReVEAL-Qwen3VL-8B on Ref-DAVIS.}
  \label{fig:layer_distribution_davis}
\end{figure}

\paragraph{VET Fusion Strategies on Qwen2.5VL-7B.}

Table~\ref{tab:tokencomb_qwen25} extends the VET fusion ablation to Qwen2.5VL-7B. Unlike Qwen3VL-8B, this backbone benefits most from background suppression, which improves attention-derived and SAM2 mask $\mathcal{J}\&\mathcal{F}$ by 4.6\% and 1.9\% on ReasonVOS, respectively. Fusing all three VET types achieves the best overall performance on both benchmarks, supporting the effectiveness of our role design and fusion strategy across MLLM backbones.

\begin{table*}[t!]
\centering
\begin{minipage}[t]{0.36\linewidth}
\centering
\centering
\setlength{\tabcolsep}{4pt}
\caption{
$\lambda_{\mathrm{CAGL}}$ and $\lambda_{\mathrm{rep}}$ ablation.}
\label{tab:cagl_weight}
\vspace{0.5mm}
{\small
\begin{tabular}{cc cc}
\toprule
\multirow{2}{*}{$\lambda_{\mathrm{CAGL}}$} &
\multirow{2}{*}{$\lambda_{\mathrm{rep}}$} &
\multicolumn{2}{c}{SAM2 $\mathcal{J}$\&$\mathcal{F}$} \\
\cmidrule(lr){3-4}
& & Ref-D & ReasV \\
\midrule
\midrule
0.1  & 0.1  & 79.4 & 62.5 \\
\rowcolor{myblue} \textbf{0.2} & \textbf{0.1} & \textbf{80.7} & \textbf{66.1} \\
0.3  & 0.1  & 79.0 & 65.9 \\
0.2  & 0.05 & 76.8 & 65.3 \\
0.2  & 0.15 & 80.3 & 66.0 \\
\bottomrule
\end{tabular}
}

\end{minipage}
\hfill
\begin{minipage}[t]{0.62\linewidth}
\centering
\centering
\setlength{\tabcolsep}{4pt}
\caption{Inference-time ablation of different VET combination strategies using
ReVEAL-Qwen2.5VL-7B. ReVEAL uses all three types of VETs during evaluation. }
\label{tab:tokencomb_qwen25}
\vspace{0.5mm}
\resizebox{\linewidth}{!}{
\begin{tabular}{l ccc cc cc}
\toprule
\multirow{2}{*}{MLLM} &
\multirow{2}{*}{\texttt{<OBJ>}} &
\multirow{2}{*}{\texttt{<BG>}} &
\multirow{2}{*}{\texttt{<NEG>}}  &
\multicolumn{2}{c}{Attn Mask} &
\multicolumn{2}{c}{SAM2 Mask} \\
\cmidrule(lr){5-6}\cmidrule(lr){7-8}
& & & & Ref-D & ReasV & Ref-D & ReasV \\
\midrule
\midrule
\multirow{4}{*}{Qwen2.5-VL-7B} & \checkmark & -- & -- & 24.8 & 17.2 & 77.3 & 62.2 \\
& \checkmark & -- & \checkmark & 25.6 & 17.8 & 77.0 & 63.5 \\
& \checkmark & \checkmark & -- & 28.6 & 21.8 & 77.4 & 64.1 \\
\rowcolor{myblue} & \checkmark & \checkmark & \checkmark & \textbf{29.4} & \textbf{22.1} & \textbf{77.7} & \textbf{65.1} \\
\bottomrule
\end{tabular}
}

\end{minipage}
\end{table*}

\paragraph{DALR Top-$k$ Ablation.}
We evaluate DALR with different top-$k$ settings, each used during both training and inference. As shown in Table~\ref{tab:router_topk}, selecting $k=6$ layers achieves the best overall performance across the two benchmarks. Selecting more layers provides no consistent gains and may introduce additional noise into the grounding evidence.

\paragraph{DALR Selection Distribution.}
Figure~\ref{fig:layer_distribution_davis} shows DALR's layer selection distribution on Ref-DAVIS with $k=6$. The selected layer combinations vary across inputs, indicating input-dependent selection rather than reliance on a fixed subset.

\paragraph{Sensitivity to the Background Suppression Weight.}

\begin{figure}[ht]
  \centering
  \includegraphics[width=1.0\linewidth]{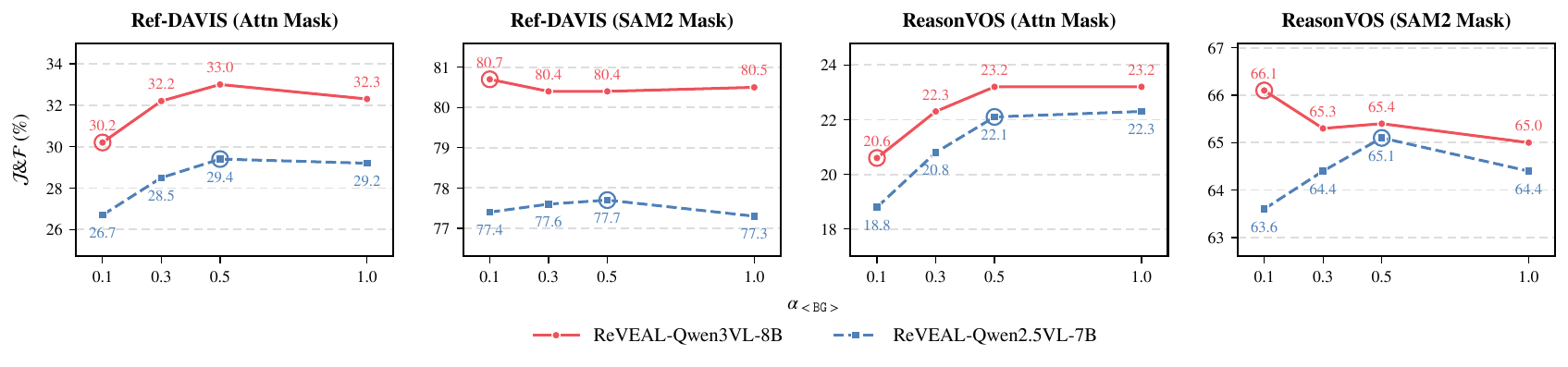}
  \setlength{\abovecaptionskip}{-12pt}
  \caption{Effect of background suppression weight $\alpha_{\texttt{<BG>}}$ across two backbones.}
  \label{fig:bg_weight}
\end{figure}

Figure~\ref{fig:bg_weight} analyzes the effect of the background suppression weight $\alpha_{\texttt{<BG>}}$ on attention-derived and SAM2 mask quality. Since \texttt{<BG>} attention often covers broad non- foreground regions, it helps suppress diffuse or spurious object attention. The required suppression strength varies across backbones due to differences in their object attention patterns. ReVEAL-Qwen3VL-8B and ReVEAL-Qwen2.5VL-7B achieve the best SAM2 performance at $\alpha_{\texttt{<BG>}}=0.1$ and $0.5$, respectively. While stronger suppression generally improves attention-derived masks, overly large weights can also remove useful target evidence and degrade final segmentation.

\paragraph{Number of Sampled Point Prompts.}

Table~\ref{tab:max_pts} studies the maximum number of points sampled from the fused attention map as visual prompts for SAM2 at inference time. Across all three benchmarks, increasing the number of sampled points yields no consistent performance gains. The default setting provides the best overall trade-off between mask quality and prompt efficiency.


\begin{table*}[t!]
\centering
\setlength{\parskip}{0pt}
\setlength{\abovecaptionskip}{0pt}
\setlength{\belowcaptionskip}{0pt}
\begin{minipage}[t]{0.48\linewidth}
\vspace{0pt}
\caption{DALR top-$k$ ablation.}
\label{tab:router_topk}
\vspace{0pt}
\end{minipage}\hfill
\begin{minipage}[t]{0.48\linewidth}
\vspace{0pt}
\caption{Maximum points ablation.}
\label{tab:max_pts}
\vspace{0pt}
\end{minipage}\par
\vspace{3pt}
\nointerlineskip
\begin{minipage}[t]{0.48\linewidth}
\vspace{0pt}
\centering
\centering
\setlength{\tabcolsep}{4pt}
{\small
\begin{tabular}[t]{l cc}
\toprule
\multirow{2}{*}{Router top-$k$} &
\multicolumn{2}{c}{SAM2 $\mathcal{J}$\&$\mathcal{F}$} \\
\cmidrule(lr){2-3}
& Ref-D & ReasV \\
\midrule
\midrule
$k=4$ & 78.4 & 65.9 \\
\rowcolor{myblue} $\mathbf{k=6}$ & \textbf{80.7} & 66.1 \\
$k=8$ & 79.5 & 64.7 \\
$k=10$ & 80.1 & 64.2 \\
$k=12$ & 79.7 & \textbf{66.6} \\
\bottomrule
\end{tabular}
}

\end{minipage}\hfill
\begin{minipage}[t]{0.48\linewidth}
\vspace{0pt}
\centering
\centering
\setlength{\tabcolsep}{4pt}
{\small
\begin{tabular}[t]{c ccc}
\toprule
Max Points & Ref-D & ReasV & Ref-YT \\
\midrule
\midrule
2  & \textbf{81.0} & 65.9 & 71.3 \\
\rowcolor{myblue} \textbf{3}  & 80.7 & \textbf{66.1} & \textbf{71.3} \\
5  & 80.6 & 66.0 & 71.3 \\
10 & 80.6 & \textbf{66.1} & 71.3 \\
\bottomrule
\end{tabular}
}

\end{minipage}
\end{table*}

\section{More Visualizations}
\label{app:visualizations}

This section provides additional qualitative results of ReVEAL, including comparisons with other methods, more role-specific visualizations, and failure cases.

\subsection{More Comparisons with Other Methods}
\label{app:vis:decaf}

\begin{figure}[h]
  \centering
  \setlength{\abovecaptionskip}{-2pt}
  \includegraphics[width=\linewidth]{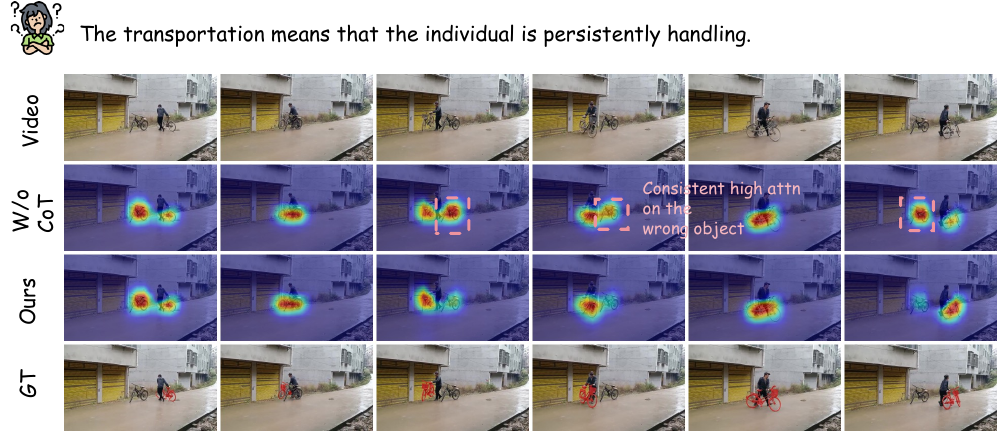}
  \caption{Qualitative comparison between ReVEAL and the Without-CoT paradigm
  on ReasonVOS.}
  \label{fig:app_noncot}
\end{figure}

Figure~\ref{fig:app_noncot} compares a representative case from ReasonVOS between ReVEAL and the Without-CoT paradigm.
In this example, ReVEAL improves attention-mask $\mathcal{J}\&\mathcal{F}$ by only 1.7\% over Without-CoT, yet its VET attention consistently favors the true target despite some hallucinated responses, whereas Without-CoT favors a visually similar distractor. Attention-mask evaluation reflects the spatiotemporal coverage of regions retained after thresholding, whereas SAM2 uses point prompts sampled from high-confidence attention locations. A higher attention-mask score does not necessarily indicate more confident attention peaks on the target. Stronger attention peaks on the true target may therefore yield more reliable point prompts, helping explain the larger gain in SAM2 segmentation despite the modest improvement in attention-mask quality.

\begin{figure}[h]
  \centering
  \setlength{\abovecaptionskip}{-2pt}
  \includegraphics[width=\linewidth]{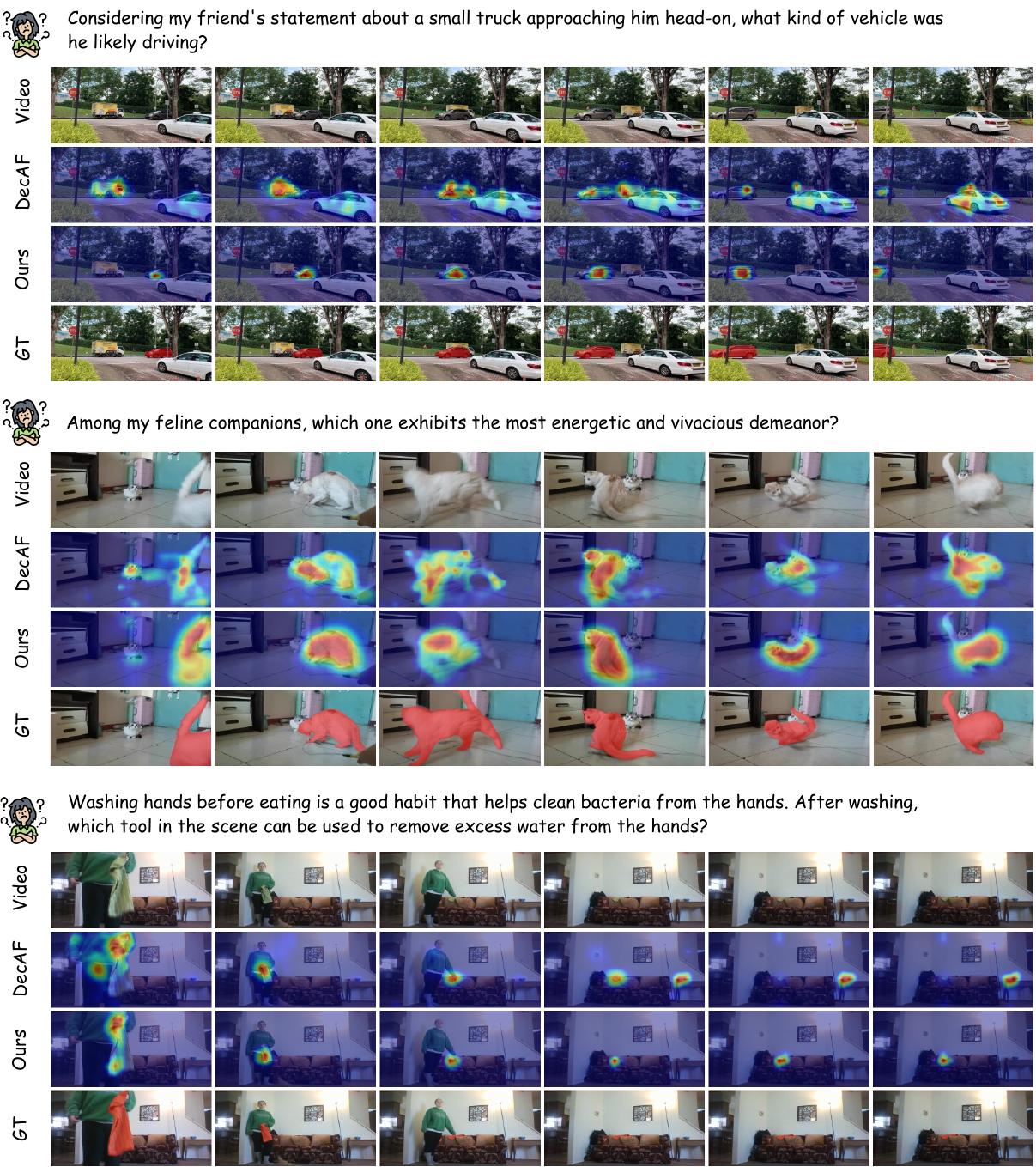}
  \caption{Additional qualitative comparisons between ReVEAL and
  DecAF.}
  \label{fig:app_decaf}
\end{figure}

Figure~\ref{fig:app_decaf} presents additional qualitative comparisons of the fused attention maps produced by ReVEAL and DecAF~\citep{decaf}. In dynamic scenes requiring target tracking across frames, ReVEAL maintains consistent and clear attention on the target, whereas DecAF tends to produce spurious activations and lose focus on the target in some cases.

\subsection{More Role-Token Attention Map Visualizations}
\label{app:vis:roles}



\begin{figure}[h]
  \centering
  \setlength{\abovecaptionskip}{-2pt}
  \includegraphics[width=\linewidth]{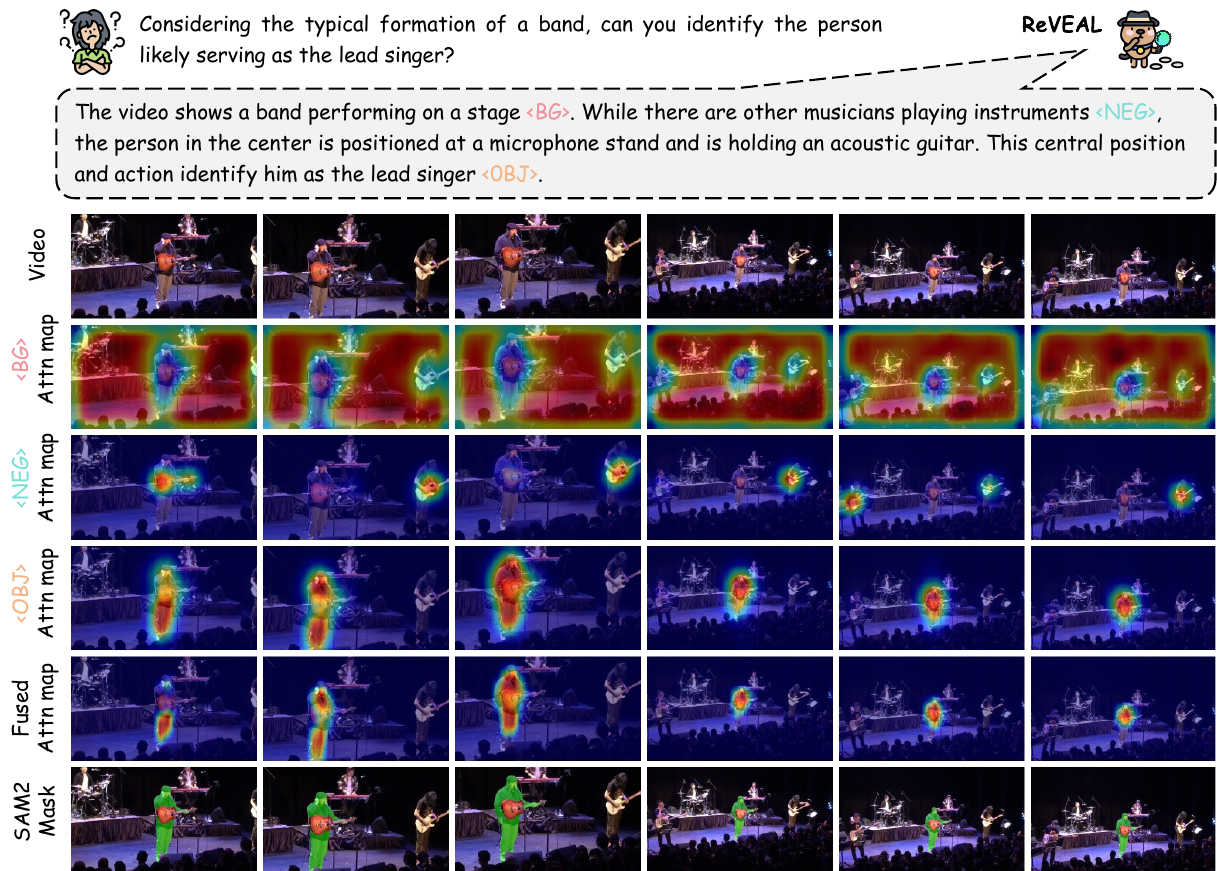}
  \caption{Additional qualitative result where all three VET roles are generated. }
  \label{fig:app_roles_music}
\end{figure}
\begin{figure}[h]
  \centering
  \setlength{\abovecaptionskip}{-2pt}
  \includegraphics[width=\linewidth]{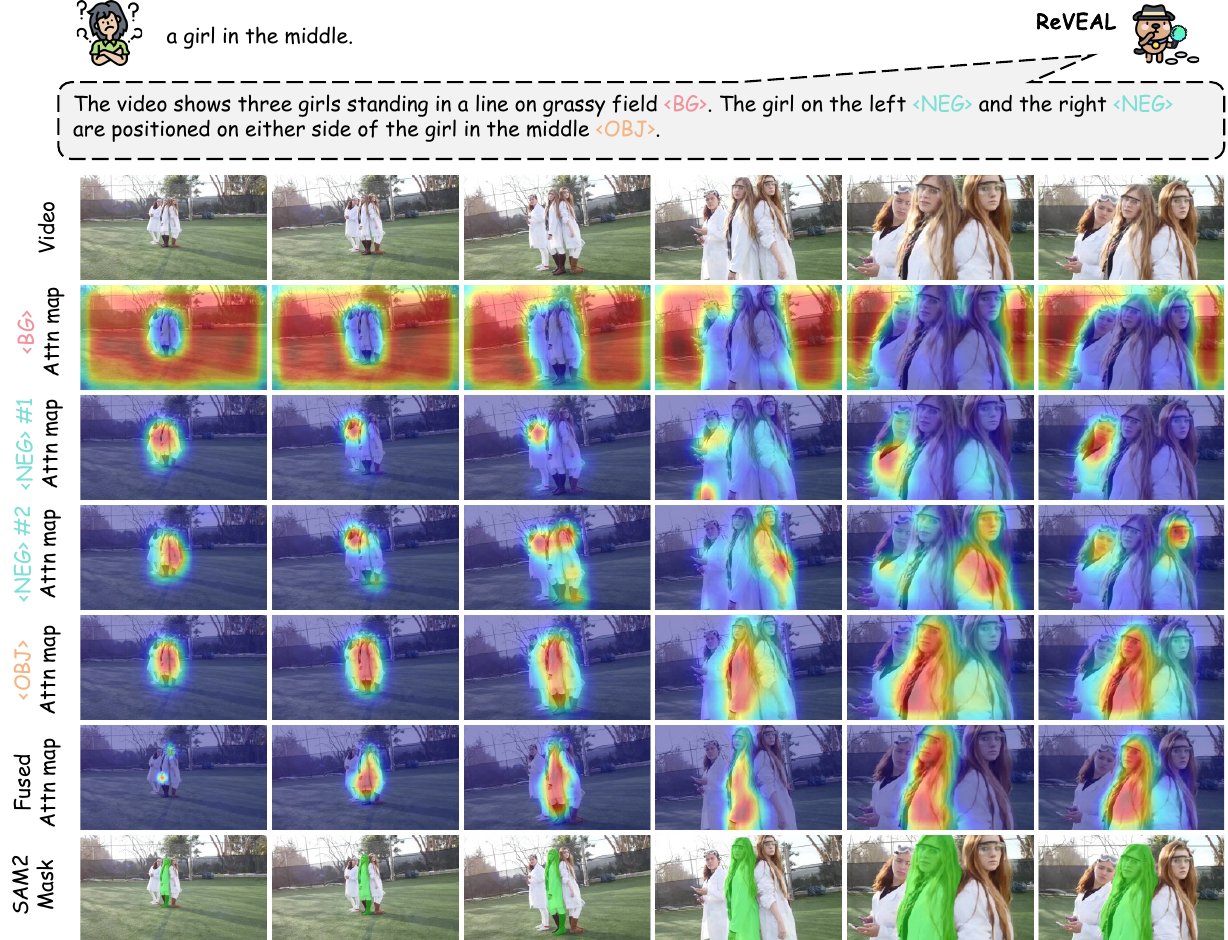}
  \caption{Qualitative result where multiple \texttt{<NEG>} tokens are generated to indicate different distactors.}
  \label{fig:app_roles_lab}
\end{figure}

Figure~\ref{fig:app_roles_music}, Figure~\ref{fig:app_roles_lab}, and Figure~\ref{fig:app_roles_milk} provide additional CoT responses and attention maps for the role-specific VETs, following Fig.~~\ref{fig:token_attn}.
ReVEAL dynamically uses suppressive VETs (\texttt{<BG>} and \texttt{<NEG>}) according to the reasoning content, with their presence and number varying across examples. The CoT reasoning itself establishes semantic associations between visual entities and their corresponding VETs, enabling individual VETs to attend to relevant regions to some extent. This lightweight association can reduce the reliance on precise region-level supervision for every \texttt{<NEG>} occurrence, which would otherwise incur additional annotation cost. As shown in Figure~\ref{fig:app_roles_lab} and \ref{fig:app_roles_milk}, different \texttt{<NEG>} occurrences tend to highlight distinct distractor regions.

\begin{figure}[h]
  \centering
  \setlength{\abovecaptionskip}{-2pt}
  \includegraphics[width=\linewidth]{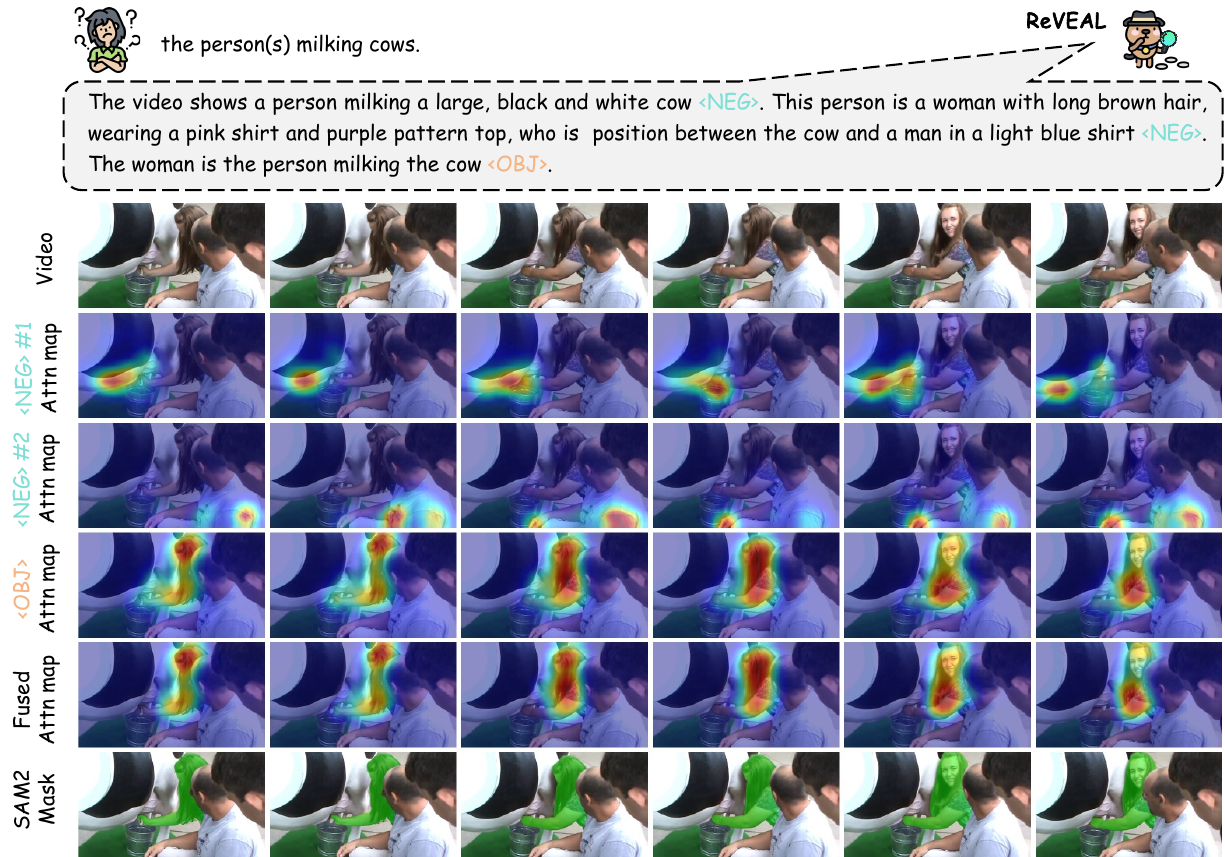}
  \caption{Qualitative result where only two type of VET roles are generated.}
  \label{fig:app_roles_milk}
\end{figure}

\subsection{Failure Case Analysis and Limitations}
\label{app:vis:failure}
\label{sec:limitations}

\paragraph{Failure cases.}

\begin{figure}[h]
  \centering
  \setlength{\abovecaptionskip}{-2pt}
  \includegraphics[width=\linewidth]{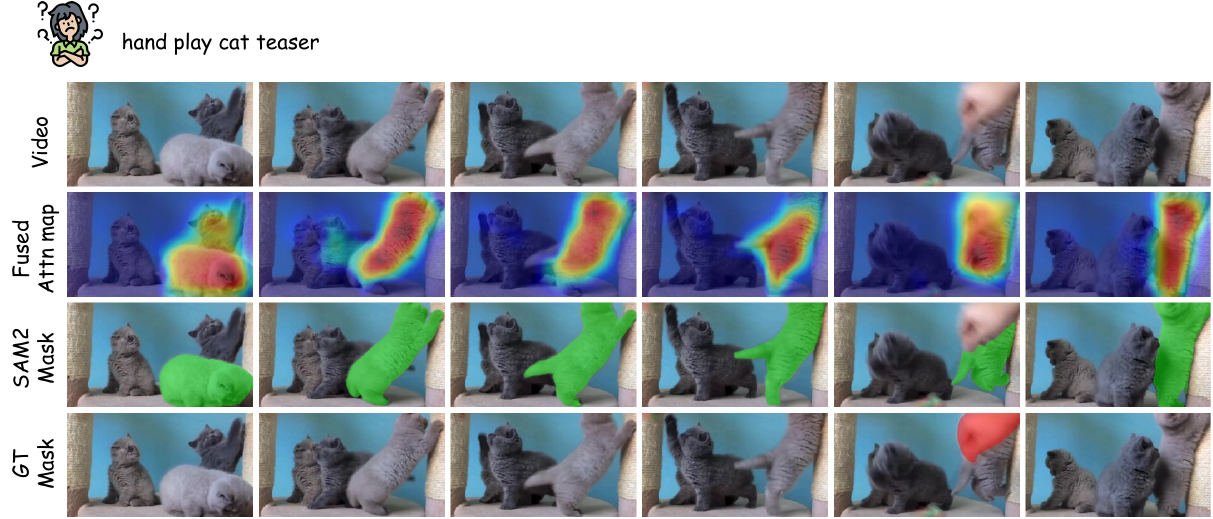}
  \caption{Representative failure case on MeViS\_u, where the fast-moving hand appears in too few of the six uniformly sampled input frames to provide sufficient temporal and visual context.}
  \label{fig:app_fc_1}
\end{figure}
\begin{figure}[h]
  \centering
  \setlength{\abovecaptionskip}{-2pt}
  \includegraphics[width=\linewidth]{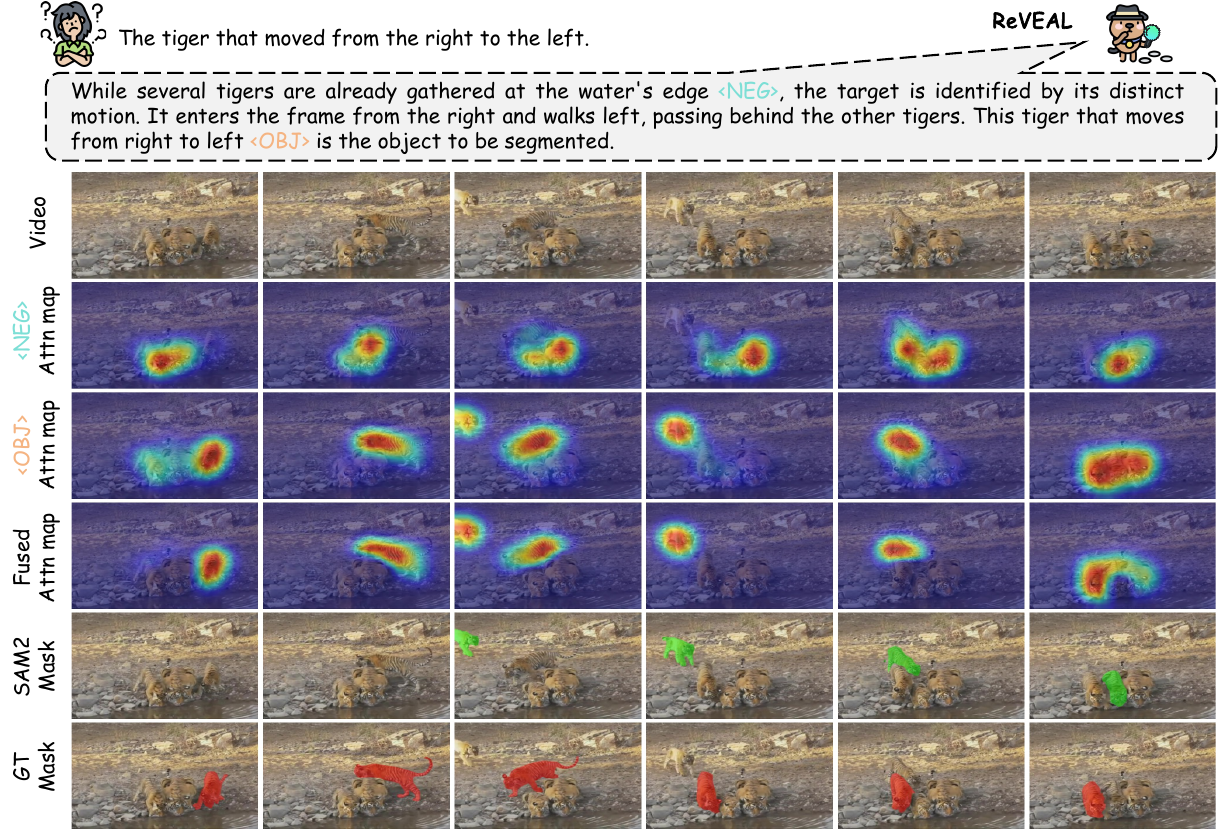}
  \caption{Representative failure case where the VET attention maps fail to distinguish the target among a group of similar tigers.}
  \label{fig:app_fc_2}
\end{figure}
\begin{figure}[h]
  \centering
  \setlength{\abovecaptionskip}{-2pt}
  \includegraphics[width=\linewidth]{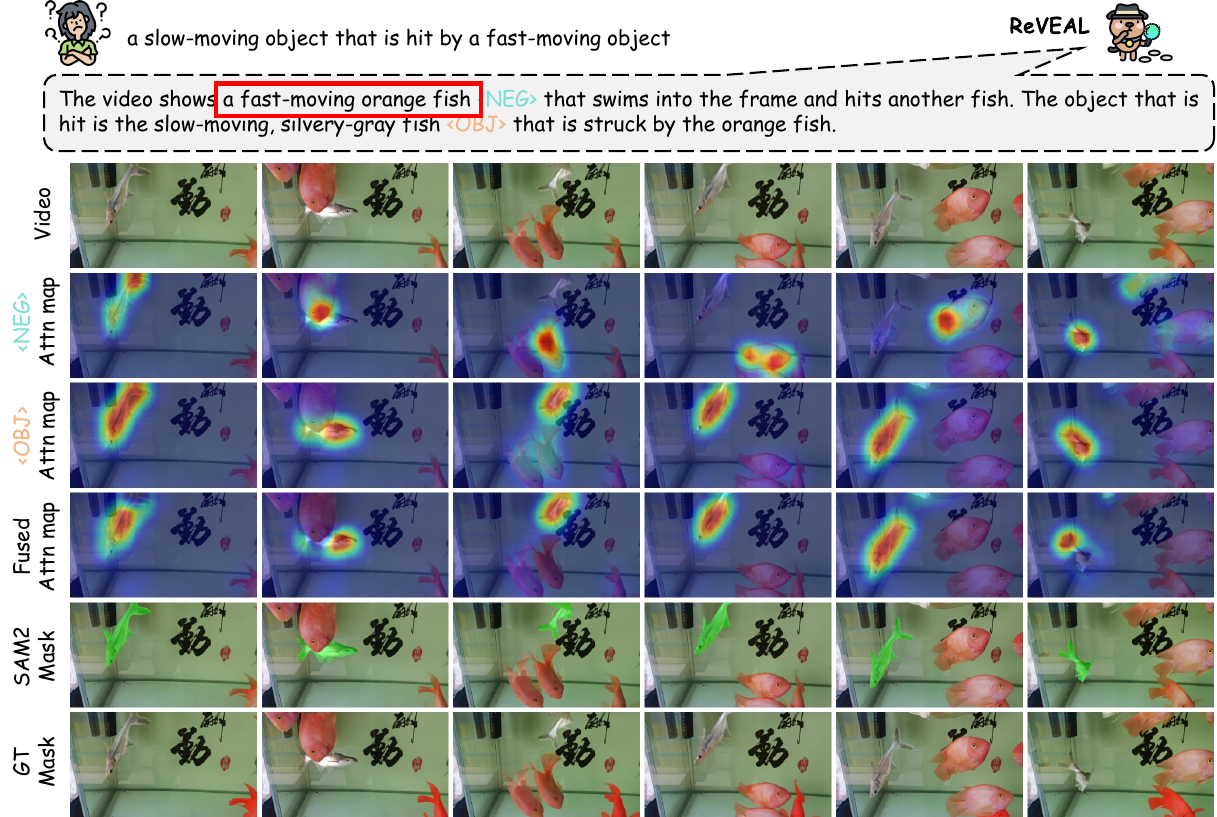}
  \caption{Representative failure case where ReVEAL fails to distinguish the target from textual reasoning when numerous visually similar distractors are present.}
  \label{fig:app_fc_3}
\end{figure}

Figure~\ref{fig:app_fc_1}, \ref{fig:app_fc_2}, and~\ref{fig:app_fc_3} show representative failures caused by limited target visibility and instance-level ambiguity. When the target is fully occluded, has not yet appeared, or is visible only briefly, limited visual and temporal context may lead to spurious attention activations in irrelevant regions. When multiple visually similar objects are present, the MLLM may struggle to identify the referred target during CoT reasoning, causing VET attention to focus on incorrect instances.


\makeatletter
\setlength{\@fptop}{0pt}
\makeatother

\paragraph{Limitations.}

Despite its effectiveness, ReVEAL has several limitations.
First, the current \texttt{<NEG>} design may not adequately distinguish individual distractors when multiple visually similar objects appear in  more complex scenes. 
Second, VET grounding can become unreliable when MLLM attention is diffuse or focused on irrelevant regions, particularly under severe occlusion.
Third, the current uniform sampling strategy may miss frames containing key visual entities, leaving them absent from the model input. The resulting lack of visual context may limit the model's ability to reason effectively and produce reliable latent visual evidence. 
Future work could explore lightweight optimizations centered on the MLLM's own video reasoning for ReVOS. For example, a frame selection module inspired by GLUS~\citep{glus} and CoT-RVS~\citep{kao2025cot} could help retain informative frames. More broadly, we will explore extending this efficient MLLM-centric design to omni scenarios.

\end{document}